%% file: main.tex
\documentclass{article}

\usepackage{microtype}
\usepackage{graphicx}
\usepackage{subcaption}
\usepackage{booktabs} % for professional tables
\usepackage{defs}
\usepackage{hyperref}
\usepackage[dvipsnames]{xcolor}

\usepackage{defs}
\usepackage{lipsum}
\usepackage{amsmath, amsthm, amssymb}
\usepackage[font=small,labelfont=bf]{caption}
\usepackage[colorinlistoftodos,textsize=tiny]{todonotes}
\usepackage{xargs}  
\usepackage{xcolor}
\newcommandx{\sjn}[2][1=]{\todo[linecolor=red,backgroundcolor=red!5,bordercolor=red,#1]{#2}}
\newcommandx{\jindong}[2][1=]{\todo[linecolor=blue,backgroundcolor=blue!5,bordercolor=blue,#1]{#2}}
\newcommandx{\zhixuan}[2][1=]{\todo[linecolor=green,backgroundcolor=green!5,bordercolor=green,#1]{#2}}
\newcommandx{\yifu}[2][1=]{\todo[linecolor=orange,backgroundcolor=orange!5,bordercolor=orange,#1]{#2}}

\usepackage[accepted]{arxiv_style}

\icmltitlerunning{Improving Generative Imagination in Object-Centric World Models}

\begin{document}

\twocolumn[
\icmltitle{Improving Generative Imagination in Object-Centric World Models}
% \icmltitle{Simulating the World via Generative Structured World Models}



\icmlsetsymbol{visit}{*}

\begin{icmlauthorlist}
\icmlauthor{Zhixuan Lin}{rucs,zu,visit}
\icmlauthor{Yi-Fu Wu}{rucs}
\icmlauthor{Skand Peri}{rucs}
\icmlauthor{Bofeng Fu}{rucs,tu,visit}
\icmlauthor{Jindong Jiang}{rucs}
\icmlauthor{Sungjin Ahn}{rucs,ruccs}
\end{icmlauthorlist}

\icmlaffiliation{zu}{Zhejiang University}
\icmlaffiliation{tu}{Tianjin University}
\icmlaffiliation{rucs}{Rutgers University}
\icmlaffiliation{ruccs}{Rutgers Center for Cognitive Science}

\icmlcorrespondingauthor{Zhixuan Lin}{zxlin.cs@gmail.com}
\icmlcorrespondingauthor{Sungjin Ahn}{sjn.ahn@gmail.com}

\icmlkeywords{Machine Learning, ICML}

\vskip 0.3in
]



% \printAffiliationsAndNotice{\icmlEqualContribution, \icmlVisitContribution} % otherwise use the standard text.
\newcommand{\icmlVisitContribution}{\textsuperscript{*}Work done while visiting Rutgers University }
\printAffiliationsAndNotice{\icmlVisitContribution} % otherwise use the standard text.

\input{00_abstract}
\input{01_intro}
\input{02_gswm}
\input{03_related_work}
\input{04_experiments}

\input{05_conclusion}

% \newpage
\section*{Acknowledgements}
SA thanks Kakao Brain and Center for Super Intelligence (CSI) for their support.

% \textbf{Do not} include acknowledgements in the initial version of
% the paper submitted for blind review.

% If a paper is accepted, the final camera-ready version can (and
% probably should) include acknowledgements. In this case, please
% place such acknowledgements in an unnumbered section at the
% end of the paper. Typically, this will include thanks to reviewers
% who gave useful comments, to colleagues who contributed to the ideas,
% and to funding agencies and corporate sponsors that provided financial


\bibliography{refs,refs_ahn}
\bibliographystyle{icml2020}

\input{supplementary}

\end{document}

%% file: 00_abstract.tex
\begin{abstract}
The remarkable recent advances in object-centric generative world models raise a few questions. First, while many of the recent achievements are indispensable for making a general and versatile world model, it is quite unclear how these ingredients can be integrated into a unified framework. Second, despite using generative objectives, abilities for object detection and tracking are mainly investigated, leaving the crucial ability of temporal imagination largely under question.~Third, a few key abilities for more faithful temporal imagination such as multimodal uncertainty and situation-awareness are missing.~In this paper, we introduce Generative Structured World Models (G-SWM). The G-SWM achieves the versatile world modeling not only by unifying the key properties of previous models in a principled framework but also by achieving two crucial new abilities, multimodal uncertainty and situation-awareness.~Our thorough investigation on the temporal generation ability in comparison to the previous models demonstrates that G-SWM achieves the versatility with the best or comparable performance for all experiment settings including a few complex settings that have not been tested before. \url{https://sites.google.com/view/gswm}
\end{abstract}

%% file: 01_intro.tex
\section{Introduction}
% Learning unsupervised world models.

Endowing machines with the ability to learn world models without supervision is a grand challenge toward human-like AI~\citep{johnson1983mental,worldmodels}.~A promising approach to this is to develop structured generative models in which world states are represented by composition of abstract entities such as objects and agents, and their relationships. This abstraction grounded in the physical properties~\citep{spelke1993gestalt} can help enable interpretability, reasoning, and optimal decision-making via efficient simulation (a.k.a., imagination) of the possible futures---a crucial ability for model-based decision making in both humans~\citep{schacter2012future,addis2012hippocampus} and AI systems~\citep{alphagozero}.

In this direction, there have been remarkable advances recently in approaches using spatial attention for unsupervised object detection, representation, and generation. These include the object propagation-discovery model in SQAIR~\citep{sqair}, the physical interaction models in STOVE~\citep{stove,rnem}, the scalability models in SCALOR~\citep{scalor} and SILOT~\citep{silot}, and the background context models in SCALOR. Although each of these abilities can be an indispensable ingredients toward the future, there are also a few key limitations and challenges that should be addressed.
% toward a general-purpose world model.

% However, considering that each of these abilities is an indispensable component to achieve a general-purpose world model, a few questions are naturally raised following these advances.

\begin{table}[t]
\centering
\caption{Summary of abilities addressed in the previous models. In the table, `Uncertainty' and `Situated' stand for multimodal uncertainty and situation-awareness, respectively. }
\vskip -4mm
\label{tb:1}
\input{tables/summary_color.tex}
\vskip -0.1in
% \vspace{-1mm}
\end{table}

A primary challenge is how we can integrate the benefits of such fragmental advances into a unified model, e.g. as shown in Rainbow~\citep{rainbow}. While all of the above-mentioned properties are essential to learning a complex world model, the models are developed with their own goals in controlled and isolated settings where many abilities other than the main one are ignored  (See Table~\ref{tb:1}). For instance, although it is usual for an object to occlude some objects while interacting with another, joint modeling of these two properties have not been investigated properly.
% --- we show that all existing models fail on this simple but realistic task. 
% Similarly, all models based on the propagation-discovery model do not support physical interaction among objects. 
It is, therefore, crucial to develop a unified model while maintaining a principled modeling framework without losing existing benefits.

Another significant problem is that, despite being trained with generative objectives, in many of the previous models, the generation abilities are not well investigated. Instead, object-centric representation (via unsupervised detection) and tracking were focused as the main task in
these models~\citep{sqair, silot, scalor}. Hence, it is largely under question how well these models can actually simulate or imagine the possible futures, which is the central purpose of world models.

Lastly, a few key abilities required to enable faithful simulation quality have been missing. Among such abilities are multimodal uncertainty and situated behavior. Multimodal uncertainty is crucial in model-based decision-making as the simulation should be capable of exploring various possibilities. Although previous models contain stochastic latent variables, being based on Gaussian modeling, their ability to explore diverse scenarios is significantly limited to stochastic unimodal search. In addition, agents should show situated behavior depending on, e.g., where it is placed in the environment. However, previous models do not support such situation-aware behavior, e.g. an agent navigating through the corridor of a maze.
% Although some models~\citep{stove, rnem} achieve good physical interactive behavior, 
% They cannot model situations like 

In this paper, we propose a new generative world model, called Generative Structured World Models (G-SWM), for unsupervised learning of object-centric state representation and efficient future simulation.
\ours not only unifies the key abilities of previous models in a principled framework but also achieves multimodal uncertainty and situated behavior.
The contributions of the paper are as follows.
First, we introduce a model with a versatile propagation module that can handle occlusion of objects as well as interactions both among objects and the environment--situation.
Second, we incorporate a hierarchical object dynamics model, enabling the simulation of stochastic dynamics and multimodal behavior.
Lastly, we thoroughly investigate the generation performance of \ours and previous models through a series of extensive experiments designed to highlight the abilities and limitations of the various models.

%% file: tables/summary_color.tex
% \vskip 0.15in
\begin{center}
\begin{small}
\begin{sc}
\begin{tabular}{lcccc}
\toprule
 & SQAIR & SILOT & SCALOR & STOVE \\
\midrule
% Discovery           & ? & ? & ? & ? \\
% Occlusion           & ? & ? & ? & ? \\
% Interaction         & ? & ? & ? & ? \\
% Two Layers          & ? & ? & ? & ? \\
% Two Layers Dense    & ? & ? & ? & ? \\
Discovery       & \textcolor{ForestGreen}{\checkmark} & \textcolor{ForestGreen}{\checkmark} & \textcolor{ForestGreen}{\checkmark} & $\times$  \\ 
Interaction     & $\times$ & $\times$ & $\times$ & \textcolor{ForestGreen}{\checkmark}  \\ 
Occlusion       & \textcolor{ForestGreen}{\checkmark} & \textcolor{ForestGreen}{\checkmark} & \textcolor{ForestGreen}{\checkmark} & $\times$     \\
Scalability     & $\times$ & \textcolor{ForestGreen}{\checkmark} & \textcolor{ForestGreen}{\checkmark} & $\times$  \\
Background         & $\times$ & $\times$ & \textcolor{ForestGreen}{\checkmark} & $\times$  \\ \hline 
% SSM & X & X & X & o  \\ \hline
Uncertainty    & $\times$ & $\times$ & $\times$ & $\times$  \\
Situated        & $\times$ & $\times$ & $\times$ & $\times$ \\
\bottomrule
\end{tabular}
\end{sc}
\end{small}
\end{center}
\vskip -0.1in

%% file: 02_gswm.tex
\section{Object-Centric World Models}
Our proposed model inherits its base architecture from SCALOR~\citep{scalor} as it supports both foreground and background models (and more generally from object-centric world models based on spatial-attention~\citep{sqair,scalor,silot}). In this section, we introduce the problem setting, basic architecture shared in such models, and notations.

% As our proposed model inherits its base architecture from the object-centric temporal generative models such as SCALOR and SILOT, we first introduce the basic problem settings with our notations.
% that closely follow those used in SILOT.

\textbf{Video, Objects, and Contexts.} Consider a video $\bx = (\bx_1, \dots, \bx_T)$ of length $T$ describing a dynamic scene containing objects interacting with each other.
We consider a generative process in which each object $k$ in the scene at timestep $t$ is generated from a latent variable $\dotbz_t^k$ and everything else (i.e. non-object related) is modeled by a context latent variable $\bz_t^\ctx$.
We assume that both objects and context are dynamic.
We use $\dotbz_t$ to denote the set of object latents $\{\dotbz_t^k\}_k$ at $t$. All latent variables at frame $t$ are denoted by $\bz_t = \dotbz_t  \cup \bz_t^\ctx$.

\textbf{Object Representation.} The object latent variable $\dotbz_t^k$ usually consists of object \textit{attribute} latent variables. With some abuse of notation, we denote this by $\smash{\bz^{\attr,k}_t = \{\bz^{\al,k}_t\}_{\al \in \cA}}$ with $\cA$ the set of attribute index. Different models may use different attribute sets. In our model, we use the following attribute set $\smash{\bz^{\att,k}_t = \{\bz^{\what,k}_t, \bz^{\where,k}_t, \bz^{\pres,k}_t, \bz^{\depth,k}_t\}}$ to represent appearance, position, presence, and depth of an object, respectively. Unlike previous models, our model assumes that these attributes of an object are generated from the \textit{state} latent variable $\smash{\bz^{\state, k}_t}$. Thus, $\smash{\dotbz_t^k = \{\bz^{\attr,k}_t,  \bz^{\state, k}_t}\}$. The state latent $\smash{\bz^{\state,k}_t}$ along with an RNN connecting it across the timesteps is used to explicitly represent object dynamics and to model multimodal uncertainty.

The attribute latents $\bz^{\attr,k}_t$ represent randomness of its target object properties but may not be in a form useful for rendering.
For example, it is more efficient to update the object position by learning the  position \textit{deviation} instead of directly predicting the absolute position.
To distinguish between the attribute latent and the explicit attribute value (in an explicit form for rendering), we introduce the notation  $\smash{\bo^k_t = \{\bo^{\pres,k}_t, \bo^{\depth,k}_t, \bo^{\where,k}, \bo^{\what,k}_t \}}$ to denote explicit attributes. 
The explicit attribute is a deterministic function of the attribute latent, e.g. $\bo_t^{\where,k} = \bo_\tmo^{\where,k} + c \cdot \tanh(\bz_t^{\where,k})$.
These attributes have the same interpretation as previous related works~\citep{scalor, silot}.

\textbf{Propagation and Discovery.} 
Using the propagation-discovery model~\citep{sqair,scalor,silot}, we denote the object latents propagated from frame $t-1$ to frame $t$ by $\tilbz_t$ and object latents discovered at frame $t$ by $\barbz_t$. Thus $\dotbz_t = \tilbz_t \cup \barbz_t$. To discover new objects, each frame $\bx_t$ is divided into a grid of $H\times W$ cells, and {each cell is associated with a discovery object latent variable $\barbz^k_t$}. During inference, these latents are inferred in parallel with a fully convolutional network~\citep{fcn}. At the end of each timestep, we select a total of $K$ objects $\bo_t$ with the highest presence from the union of the discovered objects and propagated objects $\tilbo_t\cup \barbo_t$. These objects are also the ones to be propagated to the next timestep. In this work, we use fixed prior distributions for discovered latents, so we only do discovery during inference.

\textbf{Rendering.}  To render the foreground $\bmu^{\fg}_t$, we first reconstruct individual objects using $\bo^{\what}_t$ and combine them with $\bo^{\where}_t$, $\bo^{\pres}_t$, and $\bo^{\depth}_t$. A foreground mask $\bal_t$ combining individual object masks is also obtained. The background $\bmu^{\bgr}_t$ is decoded from the context latent $\bz^{\ctx}_t$. The final reconstruction is $\bmu_t = \bmu^{\fg}_t + (1 - \bal_t)\bmu^{\bgr}_t$, and the likelihood function $p_\ta(\bx_t|\dotbz_{\le t}, \bz^{\ctx}_{\le t})$ is a pixel-wise Gaussian distribution $\mathcal N(\bx_t| \bmu_t, \sigma^2\bI)$ where $\sigma$ is a hyperparameter.

\begin{figure*}[t!]
    \centering
    \includegraphics[width=0.8\linewidth]{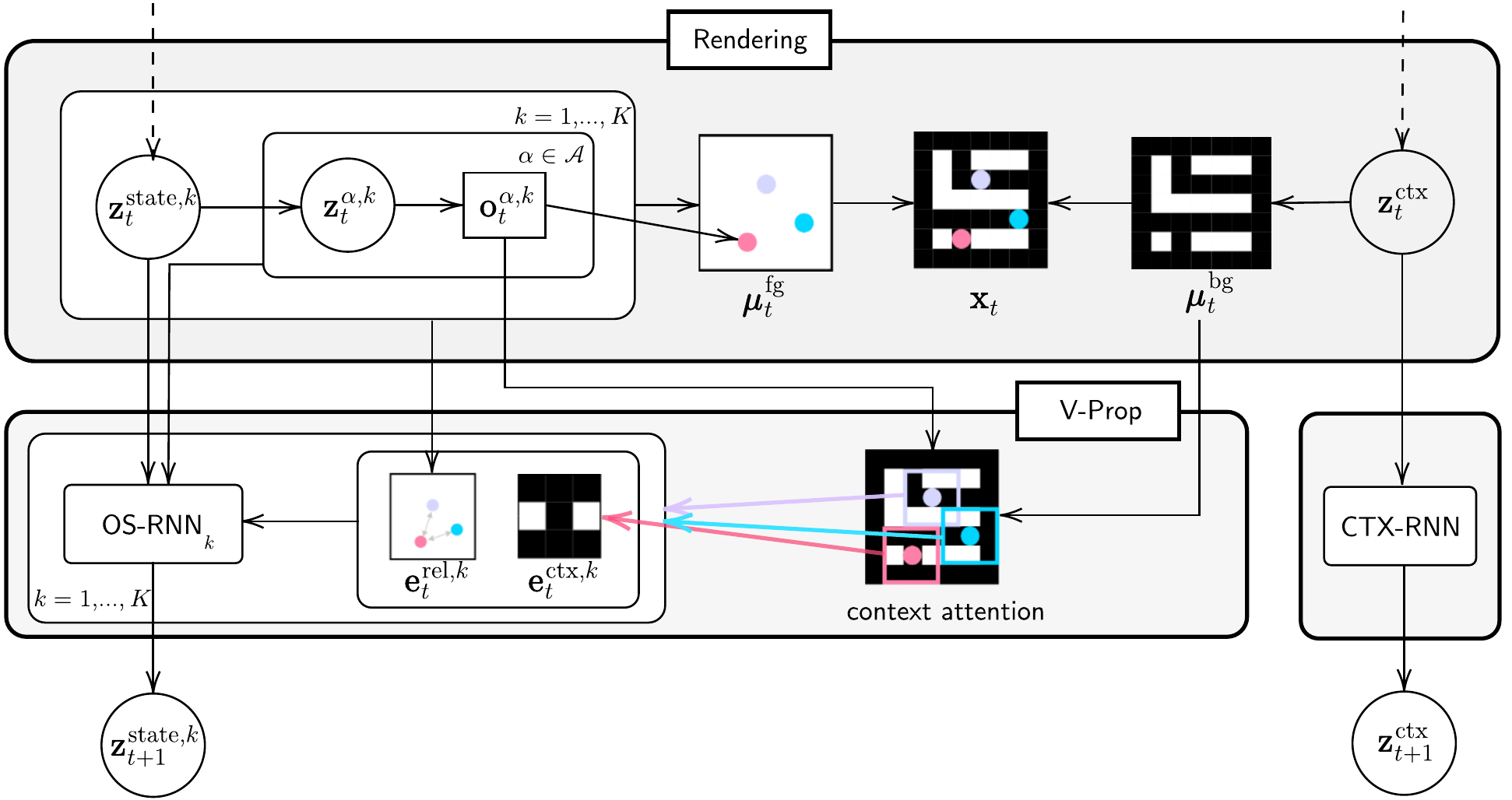}
    \caption{G-SWM Generative Process. For each timestep $t$, the generated image is a combination of a foreground image $\bmu^{\fg}$ and a background image $\bmu^{\bgr}$. To obtain the foreground image $\bmu^\fg$, we determine the object attributes by decoding $\bz^{\state,k}_t$ and render them into a canvas. The background image is decoded from the context latent variable $\bz^{\ctx}_t$. To propagate existing objects to the next timestep, the V-Prop module integrates information of object-object and object-context interaction in the object-state RNN, from which we compute the new object states $\bz^{\state, k}_{t+1}$. The context latent variable is also updated to $\bz^{\ctx}_{t+1}$ with a context RNN.}
    \label{fig:gswm}
\end{figure*}

\section{Generative Structured World Models}

In this section, we describe our proposed model, Generative Structured World Models (G-SWM). We first describe the probabilistic formulation  of the  model. Then, we describe the propagation module which contains the main contribution of the proposed model. Finally, we describe the inference and learning method.

\subsection{Probabilistic Modeling}

As a temporal generative model, G-SWM consists of the following four generation modules: (i) the \textit{context} module generates a new context representation from its history, (ii) the \textit{propagation} module updates the attributes of currently existing objects for the next timestep, (iii) the \textit{discovery} module generates new objects, and (iv) the \textit{rendering} module renders objects and the background into a canvas to generate the target image. The generative process of \ours is illustrated in Figure~\ref{fig:gswm}. 

Specifically, the generative process of a video $\bx$ of length $T$ can be described by the following joint distribution:
\eq{
    p_\ta(\bx_\ott,\bz_\ott) = \pd{t}{T}p_\ta(\bx_t, \barbz_t, \tilbz_t, \bz_t^\ctx | \bz_{<t})
}
and the single timestep model $p_\ta(\bx_t, \barbz_t, \tilbz_t, \bz_t^\ctx | \bz_{<t})$ is further decomposed into: 
\eq{
\underbrace{p_\ta(\bx_t|\dotbz_{\leq t}, \bz_{\leq t}^\ctx)}_{\text{Rendering}} \underbrace{p(\barbz_t)}_{\text{Discovery}}
\underbrace{p_\ta(\tilbz_t| \dotbz_{<t}, \bz_{<t}^\ctx)}_{\text{Propagation}}
\underbrace{p_\ta(\bz_t^\ctx|\bz_{<t}^\ctx)}_{\text{Context}} \,.\nn
}
The main contribution of the G-SWM model is implemented in the propagation and context module. Other modules are similar to the ones proposed in SCALOR and SILOT. Thus, in the following, our description of the model focuses the context and propagation modules. For the implementation of the discovery and rendering modules, refer to Appendix.

\subsection{Context Modeling}
The context latent variable $\bz^{\ctx}_t$ represents the state of global context, i.e., the background, that may affect the dynamics of the objects as well as the appearance of the background. For generality, we assume a dynamic context. The dynamics of the context is modeled by the conditional distribution $p_\ta(\bz^\ctx_t| \bz^\ctx_{<t})$ which is a Gaussian distribution. This is implemented with an RNN followed by an Multi-layer Perceptron (MLP) yielding the parameters of the Gaussian distribution.

\subsection{Versatile Propagation} 

% \begin{figure}[tbp]
%     \centering
%     \includegraphics[width=0.8\linewidth]{figs/vprop}
%     \caption{Versatile propagation.}
%     \label{fig:prop}
% \end{figure}

The core of our model is the propagation module which we call the versatile propagation (V-Prop) module. This module integrates diverse key abilities of existing object-centric world models into a unified framework as well as implementing new abilities such as situated behaviour and multimodal uncertainty. As a result, the V-Prop module can support complex object interaction-occlusion, situated behaviour, and multimodal uncertainty jointly. In this section, we describe how these abilities are implemented in our model.

\textit{Object-State RNN.} The backbone of the V-Prop module is the object-state RNN (OS-RNN) associated with each object (Figure~\ref{fig:gswm}). The object-state RNN tracks dynamics of various states of its associated object. In particular, it takes as input (1) the interaction encoding of the object with other objects and (2) the context encoding modeling interaction with the background from the perspective of object $k$. Then, the state of the OS-RNN $\bh_t^k$ is updated as follows:
\eq{
    \bh_t^k = \RNN([\bo_\tmo^k, \bz_\tmo^{\state, k}, \bee_\tmo^{\ctx,k}, \bee_\tmo^{\rel,k}], \bh_\tmo^k)\,.
    \label{eq:ornn}
}
Here, $\bee_\tmo^{\ctx,k}$ is the \textit{object-context} interaction encoding and $\bee_\tmo^{\rel,k}$ is the \textit{object-object} interaction encoding. In the following, we describe in more detail how these representations are obtained.

% \subsubsection{Object Interaction}
% Unlike previous models that only support either physical interaction like collision or occlusion without collision, G-SWM can deal with both interaction \textit{and} occlusion in a scene by learning it from object behaviours. We implement this ability by modeling pairwise interaction between objects using a graph neural net. 

\subsubsection{Interaction \& Occlusion} 

Objects and agents in a physical environment interact with complex relations. Collision and occlusion are among two most popular types of such object interactions. Occlusion may not be seen a direct interaction between objects when the objects are physically disconnected. However, from the perspective of an observer, occlusion can still be modeled as an implicit interaction between the objects where one object makes another one invisible fully or partially. Although in a realistic environment these are the most fundamental types of object relationship that can happen simultaneously in a scene, in previous works these are studied separately and thus its jointly modeling and working have not studied.

In our model, we model both collision and occlusion in a simple graph neural networks (GNN). This object interaction modeling is also used in STOVE. However, in our model, we take the depth of the object into consideration and thus make the GNN considers not only object position but also object depth to model both the collision and occlusion. In our experiments, we found that this interaction modeling can learn different behaviours depending on the depth of an object and situation. 

Specifically, we compute the interaction encoding $\bee^{\rel, k}_{t}$ of object $k$ as follows:
\eq{
    \bee_t^{\rel,k} = \bee_t^{k,k} + \sum_{j\neq k} w_t^{k,j}\bee_t^{k,j}.
}
To perform this, we first obtain the object representation $\bu_t^k$ by concatenating object attributes $\bo_t^k$, the state latent $\bz_t^{\state, k}$, and the temporal object state $\bh_t^k$. Then, we obtain the self-interaction encoding by $\bee_t^{k,k} = \MLP(\bu_t^k)$, and the pairwise-interaction encoding $\bee_t^{k,j}$ and the interaction weight (normalized with softmax) $w_t^{k,j}$ by $(\bee_t^{k,j}, w_t^{k,j})=\MLP(\bu_t^k, \bu_t^j)$.

% $e^{\inter, k}_{t-1}$ integrates information from other objects. In this work, we consider pairwise interaction with a graph neural net as in STOVE, but more complicated interaction modules can be used. 

% \subsubsection{Situation Awareness}
\subsubsection{Situation Awareness} 
Objects not only behave interactively with other objects but also do so depending on their spatial situation. For example, an agent in PacMan should not only behave depending on the enemies but should also follow the corridor of a maze while not penetrating the wall.

We define such dependency of object states to the local spatial properties of the environment as \textit{situation-awareness}. To implement this, we model the interaction between an object $k$ and the environment by computing a context encoding $\bee_t^{\ctx,k}$ using a simple attention mechanism, \textit{attention on environment} (AOE). 

The context encoding is obtained by first cropping the local area of the object using the Spatial Transformer (ST)~\citep{spatial_transformer} and then encoding the cropped patch:
\eq{
    {\bg}_t^{\ctx,k} &= \text{ST}(\bmu_{t}^\bgr, \bo_t^{\pos,k}, s^\ctx)\ , \\ 
    \bee_t^{\ctx,k} &= \CNN({\bg}_t^{\ctx,k})\,.
}
Here, $\bmu_t^\bgr$ is the background image generated from $\bz_t^{\ctx,k}$ and $\bo_t^{\pos,k}$ is the object position. A glimpse image ${\bg}_t^{\ctx,k}$ extracted from $\bmu_t^\bgr$ using attention by a Spatial Transformer is then encoded into the situation encoding $\bee_t^{\ctx,k}$ using a CNN. The glimpse size is controlled by hyperparameter $s^\ctx$. In other words, this is an encoding of \textit{what happened nearby the object}. This situation encoding is then provided as input to the object-state RNN in Eqn.~\ref{eq:ornn} to model temporal dynamics while being mixed with other encodings.

\subsubsection{Multimodal Uncertainty} 
One of the main uses of a world model \citep{Ha2018WorldM} in model-based reinforcement learning or planning is to simulate diverse possible futures by learning a future distribution. Although previous models contain uncertainty modeling, their flexibility is limited to unimodal uncertainty because of the reliance on Gaussian distributions which provide only the unimodal uncertainty model. For instance, in predicting the future direction of an object, it would be hard to model the possibility of two very different directions, e.g. turning left or right, although it can model some randomness towards a direction, e.g. turning more or less to the left.

To this end, we adopt hierarchical latent modeling by introducing a high-level object-state latent variable $\tilbz_t^{\state,k}$ and from the state latent we generate the attribute latents. This can be modeled by $p_\ta(\tilbz_t^{\att, k}, \tilbz_t^{\state,k} | \bz_{<t}) =$
\eq{
 p_\ta(\tilbz_t^{\attr, k} | \tilbz_t^{\state,k}) p_\ta(\tilbz_t^{\state,k}|\bz_{<t})\,.
\label{eq:hlvm}
}
% Given $h^{k}_t$, the sampling process is implemented as follows:
% \begin{align*}
%     [\tmu^{\state,k}_t, \tsg^{\state,k}_t] &= \MLP^{\state}_{\prior}(h_t^k)\ ,\\{}
%     \tz^{\state,k}_t&\sim \mathcal N(\tmu^{\state,k}_t, \tsg^{\state,k}_t)\ ,\\{}
%     [\tmu^{e,k}_t, \tsg^{e,k}_t] &= \MLP^e_\prior(\tz^{\state,k}_t)\ ,\\{}
%     \tz^{e, k}_t&\sim\mathcal N(\tmu^{e,k}_t, \tsg^{e,k}_t)\ .
% \end{align*}
That is, by learning the conditional behaviour distribution $p_\ta(\tilbz_t^{\state,k}|\bz_{<t})$, we can express the distribution of the expected behaviour of an object as a multivariate Gaussian distribution. Then, conditioning on a sampled behaviour, the object attribute $\tilbz_t^{\attr, k}$, such as position, can express highly complex multimodal uncertainty. 

To model $p_\ta(\tilbz_t^{\state,k} | \bz_{<t})$, we use a multivariate Gaussian distribution whose parameters are estimated by an MLP taking $\bh_t^k$ of object-state RNN as input. Then, latent variables $\tilbz_t^{\att, k}$ required to obtain object attributes $\tilbo_t^k$ are sampled from Gaussian or Bernoulli distributions whose parameters are computed by MLPs taking $\tilbz_t^{\state,k}$ as input. Refer to Appendix for more details of the implementation.

\subsection{Inference and Learning}

\subsubsection{Inference}
For context inference, we use an approximation 
\eq{
q_\phi(\bz^\ctx_t|\bz^\ctx_{<t}, \bx_t) = \cN(\bmu^\ctx, \bsig^\ctx)
} 
which has a similar structure as the generation but with the encoding of $\bx_t$ as an additional input to the MLP generating $\bmu^\ctx$ and $\bsig^\ctx$.

For propagation inference, given the factorization in Eqn.~\ref{eq:hlvm}, it may be tempting to use the following full factorization of the approximate posterior $q_\phi(\tilbz_t^k | \bx_t, \bz_{<t}) = $
\eq{
q_\phi(\tilbz_t^{\attr, k}|  \tilbz_t^{\state,k}, \bx_t, \bz_{<t})q_\phi(\tilbz_t^{\state,k}|\bx_t, \bz_{<t})\ .
}
However, since $\bx_t$ provides all necessary information for accurately inferring the attributes $\tilbz_t^{\attr,k}$, this can easily make the model ignore $\tilbz_t^{\state,k}$ during inference, consequently failing to learn $p_\ta(\tilbz_t^{\att, k}|\tilbz_t^{\state,k})$ as well. To address this issue, we use the following approximation replacing the factor for $\tilbz_{\att, t}^k$ inference by the prior generation model. Thus, we have $q_\phi(\tilbz_t^k | \bx_t, \bz_{<t}) = $
\eq{
p_\ta(\tilbz_t^{\attr,k}| \tilbz_t^{\state,k})q_\phi(\tilbz_t^{\state,k}|\bx_t, \bz_{<t})\ .
}
Implementation of the posterior $q_\phi(\tilbz_t^k | \bx_t, \bz_{<t})$ follows a similar structure of the prior. In particular, a separate posterior object-state RNN is used to obtain the posterior hidden state $\hatbh_t^k$ as an encoding of the history $\bz_{<t}$, and an encoding of a local proposal area from which $\tilbz^\state$ is inferred is extracted from $\bx_t$.  

\subsubsection{Learning}
Due to the posterior intractability, we use variational inference to train the model by maximizing the following evidence lower bound:
\begin{equation}
    \mathcal L(\theta, \phi) = \eE_{q_\phi(\bz|\bx)}\left[\frac{p_\ta(\bx|\bz)p_\ta(\bz)}{q_\phi(\bz|\bx)}\right]\ .
    \label{elbo}
\end{equation}

We adopt the standard ELBO in Eqn.~\eqref{elbo} as the training objective. Besides, we found that including an auxiliary loss $\KL(q_\phi(\tilbz^{\pres}|\cdot) | \Bern(p))$ with a small $p$ encourages the model to remove duplicate propagation during inference. This term also helps regularize the model and can be turned off after certain steps. We also apply the rejection mechanism proposed in SCALOR and the curriculum learning and discovery dropout used in SILOT. In our experiments, we found both these tricks help stabilize the training process. Related hyperparameters are provided in Appendix.

% \textit{Curriculum Learning, Discovery Dropout, and Rejection:} 

%% file: 03_related_work.tex
%==============================================

\section{Related Work}

Our work builds on top of a line of recent research in unsupervised object-centric representation learning.
AIR~\citep{air}, SPAIR~\citep{spair}, SuPAIR~\citep{supair}, and SPACE~\citep{space} incorporate structured latent variables with a VAE~\citep{vae} in order to detect objects in a scene without any explicit supervision.
SQAIR~\citep{sqair} extends the AIR model by combining the object discovery module with a recurrent propagation module that can track changes in detected objects over time.
SCALOR~\citep{scalor} and SILOT~\citep{silot} scale the SQAIR model to work on a large number of objects, using a parallel inference mechanism similar to SPACE. SCALOR and SPACE also introduce the dynamic background model and the background decomposition model, respectively.

STOVE~\citep{stove} introduces a state-space model for videos, combining an image model with a graph neural network (GNN) based dynamics model~\citep{graphneuralnetmodel, gatedgraphsequencenn, gcn, mpnn, battaglia2018relational, santoro2017asn, sanchezgonzalez2018graphna, interactionnet, vin} that can model interactions between objects.
STOVE leverages SuPAIR~\citep{supair} as an object detection module and can achieve long-term video generation.
R-NEM~\citep{rnem} uses a spatial mixture model learned via neural expectation maximization~\citep{nem} to obtain disentangled representations for each entity in a scene and models the physical interactions between entities.
Other works in this line of research include \citep{ddpae, tba}.
Table \ref{tb:1} summarizes the abilities and limitations of a few of the most related models.

%All models except STOVE have a discovery phase that allows the introduction of new objects.
%While SILOT and SCALOR incorporate techniques that condition the propagation of an object on its surrounding area to allow for tracking of objects that interact with one another, their mechanisms fail to successfully model interactions during generation, which we demonstrate in our experiments below.
%\ours has all the capabilities of these previous models while also being able to handle multimodal uncertainty and situated behavior.

There have also been several recent works that utilize object-centric representations in the context of model-based reinforcement learning. OP3~\citep{op3} uses a spatial-mixture model similar to IODINE~\citep{iodine} to obtain a disentangled representation of entities in a scene.
Similar to \oursns, OP3 processes entities symmetrically and uses pairwise interaction modeling.
However, it does not obtain explicit object locations and assumes the latent states follow the Markov property.
COBRA~\citep{cobra} is another spatial-mixture model that uses MONet~\citep{monet} to obtain object representations, but it is not able to handle interactions between entities.
Transporter~\citep{transporter} uses KeyNet~\citep{keynet} to discover keypoints and perform long term keypoint tracking. It makes the assumption that consecutive frames only differ in objects' pose or appearance.

%% file: 04_experiments.tex
\section{Experiments}
We evaluate \ours on several datasets designed to illustrate generation quality with respect to the different abilities outlined in Table \ref{tb:1}. We also test \ours in a 3D environment similar to the CLEVRER \citep{clevrer} dataset to show how our model performs in a more realistic setting.

% \textbf{Baselines.} We compare our model with three baselines in our experiments: SCALOR, SILOT, and STOVE. For each baseline, we use the respective authors' implementation of the model with minor modifications to report consistent metrics across the different models.

% \textbf{Evaluation Metrics.} For measuring generation performance, we use the euclidean distance between the predicted center point and the ground truth center point, similar to \citet{stove}.
% For tracking, we report the Multi-Object Tracking Accuracy (MOTA) \citep{mot}, with an IoU threshold of 0.5.
% MOTA incorporates false positive, false negatives, and id swaps (when a tracked object is mistakenly swapped for a different object) into one metric.

% \begin{figure}[thbp]
\begin{figure}[t]
    \centering
    \includegraphics[width=\linewidth]{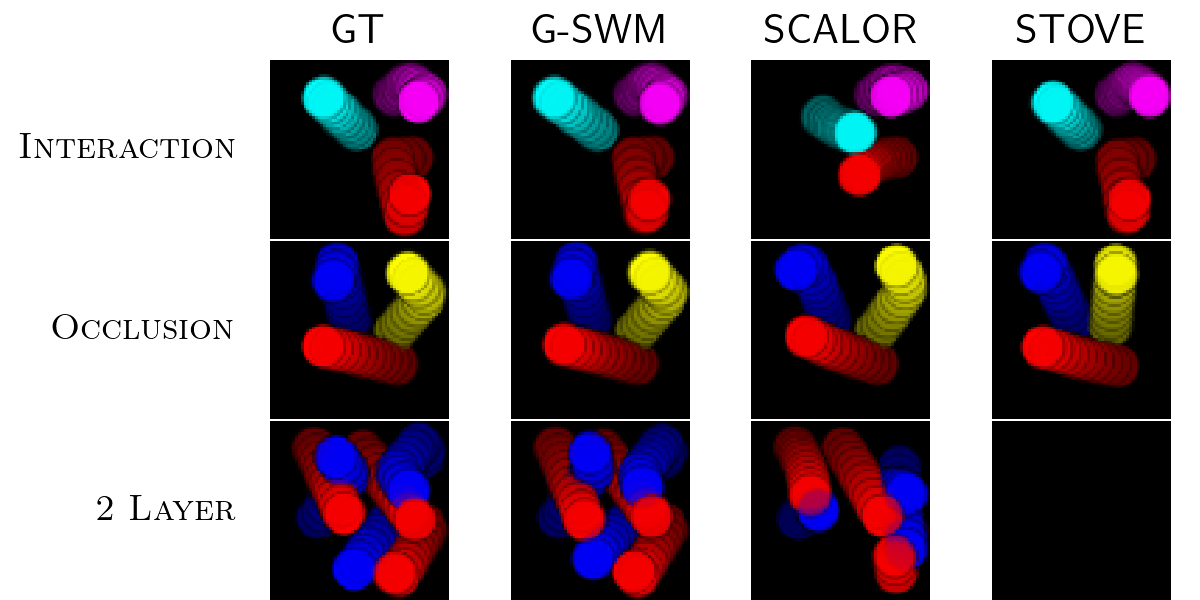}  
    \caption{Generated trajectories for the bouncing ball datasets. Conditioning on 10 ground truth frames, the first 20 generation steps are shown. To reduce visual clutter, results of the \textsc{2 Layer-D} setting are given in Appendix. 
    % {Does this occlusion result make sense?}
    }
    \label{fig:bouncing_ball}
\end{figure}

\subsection{Interaction, Occlusion, and Scalability}

In these experiments, our dataset consists of a set of balls bouncing in a frame. Four different settings are tested. In both the \textsc{Occlusion} and \textsc{Interaction} settings, there are 3 balls each with a random color. In the \textsc{Occlusion} setting the balls pass through one another according to their respective depths. In the \textsc{Interaction} setting, the balls bounce off one another. The \textsc{2 Layer} setting combines these two settings with 3 red balls at one depth and 3 blue balls at a lower depth, so the red balls occlude the blue balls. When balls of the same color touch, they bounce off one another. The \textsc{2 Layer-D} (dense) setting is the same as the \textsc{2 Layer} setting, except there are 8 balls at each depth. 
% Each episode contains 100 timesteps and we train the models on random sequences of length 20. For generation, each model is given the first 10 ground truth frames for an episode and tasked to generate a prediction of the remaining 90 frames.

% We split our data into 10,000 episodes for the training set, and 200 episodes each for the validation set and test set. Each episode is 100 timesteps and we train the models on random sequences of length 20. We use 3 random seeds to run the experiments per model per dataset.

We compare our model with SCALOR and STOVE in these experiments. Each episode contains 100 timesteps and we train the models on random sequences of length 20. For generation, each model is given the first 10 ground truth frames and asked to predict the remaining 90 frames. For measuring generation performance, we use the euclidean distance between the predicted and the ground truth center points of the balls at each timestep  \citep{stove}.

% \begin{figure}[htbp]
\begin{figure}[t]
    \centering
    \includegraphics[width=\linewidth]{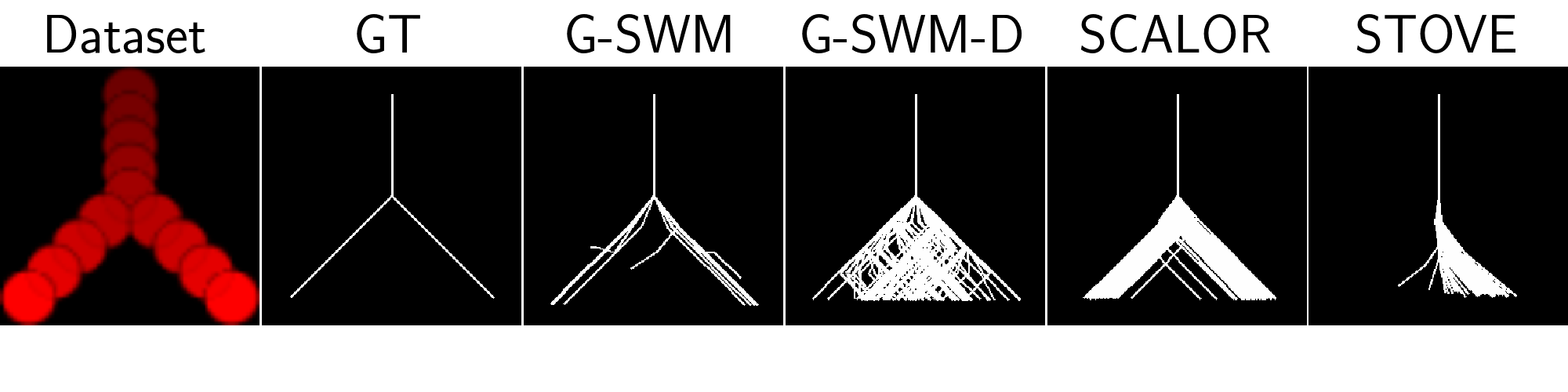}
    \caption{Generated trajectories for the the random single ball experiment. For each model, we show 100 samples.}
    \label{fig:single}
\end{figure}

% \begin{figure}[htbp]
\begin{figure}[t]
    \centering
    \includegraphics[width=\linewidth]{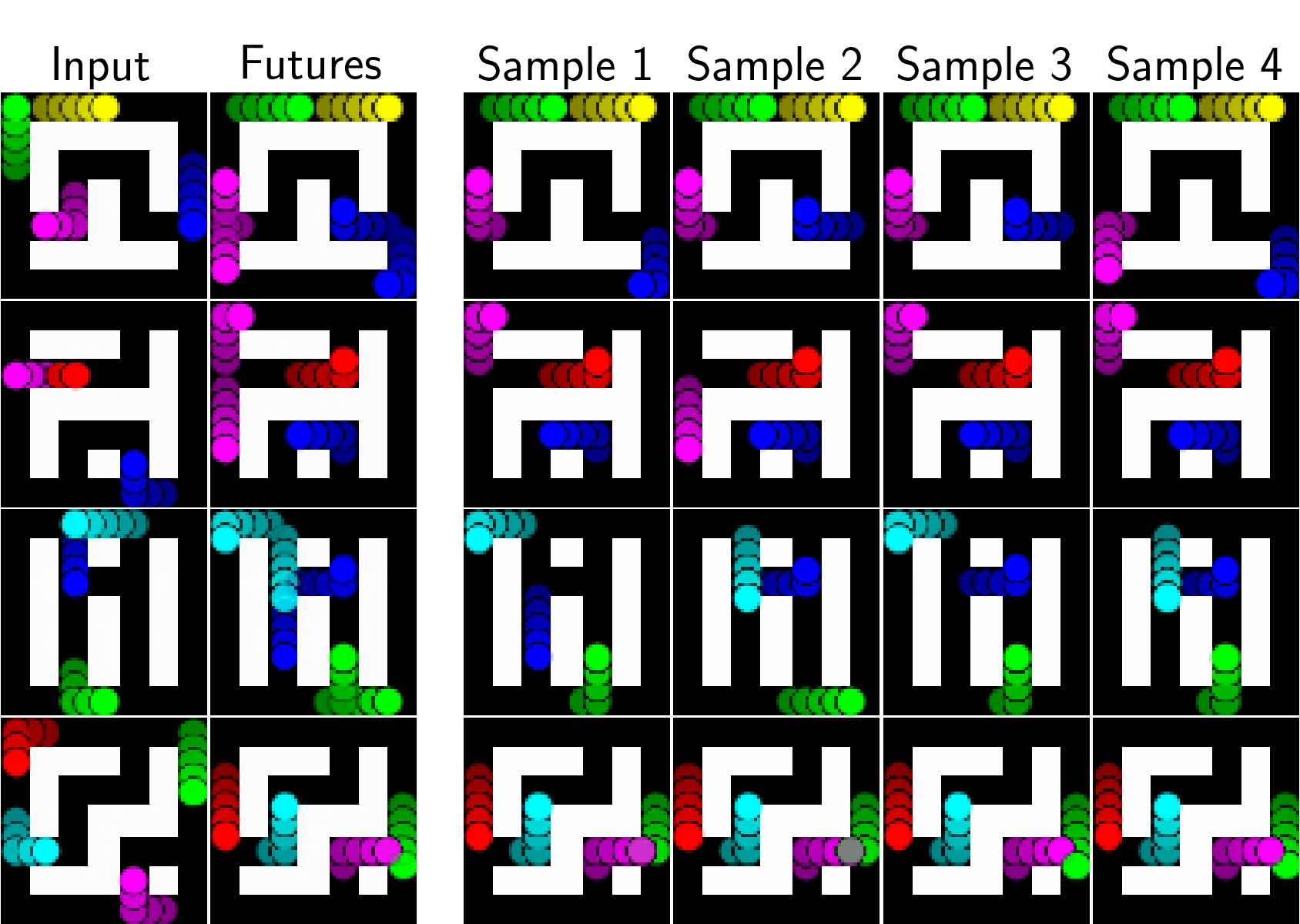}
    \caption{Generated frames for the maze experiments. The Input column shows the 5 ground truth timesteps we provide to the model. The right-hand side of the figure shows 4 possible samples given the input. All 4 samples are combined in the Futures column, representing all the predicted states of the agents. Objects stay in corridor and randomly change their directions at intersections.}
    \label{fig:maze}
\end{figure}

\begin{figure*}[t!]
\vspace{-3mm}
\centering
\includegraphics[width=\linewidth]{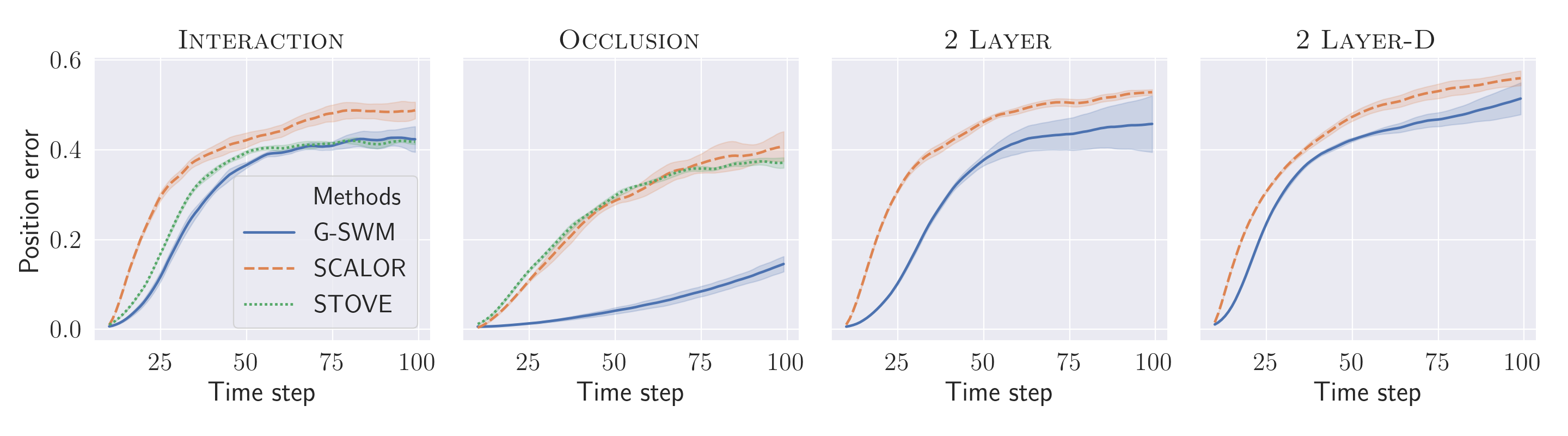}
\vspace{-8mm}
\caption{Plots of euclidean distance error of predicted ball positions per timestep. Each experiment is run with 5 different random seeds. Shaded areas indicate standard deviation.}
\label{fig:generation}
\end{figure*}

\begin{table*}[t!]
\vspace{-3mm}
\caption{Euclidean distance error of predicted ball positions, summed over the first 10 timesteps of generation.}
\vskip 0.1in
\label{generation-table}
\centering
\input{tables/all_in_one.tex}
\end{table*}

Figure~\ref{fig:bouncing_ball} shows qualitative results for the generated ball trajectories. Figure~\ref{fig:generation} plots the prediction error over 100 timesteps, and Table~\ref{generation-table} shows the summed error over the first 10 prediction steps. {STOVE performs reasonably well} in both the \textsc{Interaction} and \textsc{Occlusion} settings where there are only 3 balls, but it cannot even detect the objects in the two \textsc{2 Layer} settings, so we leave these results empty. This is because it lacks a proper mechanism for handling the frequent occlusions in the \textsc{2 Layer} settings, and its LSTM-based detection is not scalable. SCALOR can handle occlusions with $\bz^\depth$ and is scalable to the \textsc{2 Layer-D} setting, but it fails to correctly predict the ball trajectories in the \textsc{Interaction} and the two \textsc{2 Layer} settings due to lack of interaction modeling. In many cases, the balls simply slow down and stop when they collide. \ours performs the best in all settings, showing a consistent ability to handle interaction and occlusion at the same time, while being scalable to videos with a large number of objects.

\subsection{Multimodal Uncertainty and Situation Awareness}
% In this part, we evaluate G-SWM's ability to handle multimodal uncertainty and situation awareness. 
As a simple demonstration of how different models work with multimodal uncertainty, we construct a dataset with a single ball moving down the center of the frame, and then randomly changes direction and moves towards either the bottom left corner or the bottom right corner. See Figure~\ref{fig:single}. At test time, each model is provided the first several frames before the ball changes direction. Even in this simple scenario, SCALOR, STOVE, and G-SWM-D (G-SWM without $\bz^\state)$ fail to produce trajectories that match the dataset. In contrast, by introducing $\bz^{\state}$, \ours is able to model this multimodal behaviour and produce cleaner trajectories.
% Without $z^\state$, G-SWM-D's unimodal prediction is an average of the two trajectories. Similarly, STOVE's state-space model also cannot predict the correct trajectory. SCALOR, on the other hand, seems to exhibit some multimodal prediction capability, even though its architecture does not explicitly provide a mechanism to model this kind of uncertainty.
% One possible explanation for this is that SCALOR's latent variable $z_t$ at a certain timestep is directly dependent on $z_{t-1}$. In this scenario, $z_{t-1}$ may implicitly serve as an extra layer of latents, similar in function to $z^\state$, allowing for multimodal predictions.

For a more complicated scenario, we construct a maze environment (see Figure~\ref{fig:maze}) to demonstrate the ability of \ours to create realistic generations based on both interactions with the environment and multimodal uncertainty. Each episode in the dataset consists of a randomly created maze and several agents navigating through the maze. The agents only move within the corridors and continue in a straight path until they reach an intersection, at which point they choose a corridor to continue at random. For this experiment, we train on sequences of length 10 and at test time provide 5 ground truth steps to generate the following steps. 

Figure~\ref{fig:maze} shows a few examples of the generated samples. More examples are given in Appendix. These examples show that \ours is able to generate predictions where the agents correctly stay within the corridors. Additionally, the random behaviour of the agents is also accurately modeled, as the predictions show the agents moving in different valid paths for different rollouts. 

\begin{figure*}[t!]
    \centering
    \begin{subfigure}[b]{0.35\linewidth}
      \centering
      \includegraphics[width=\linewidth]{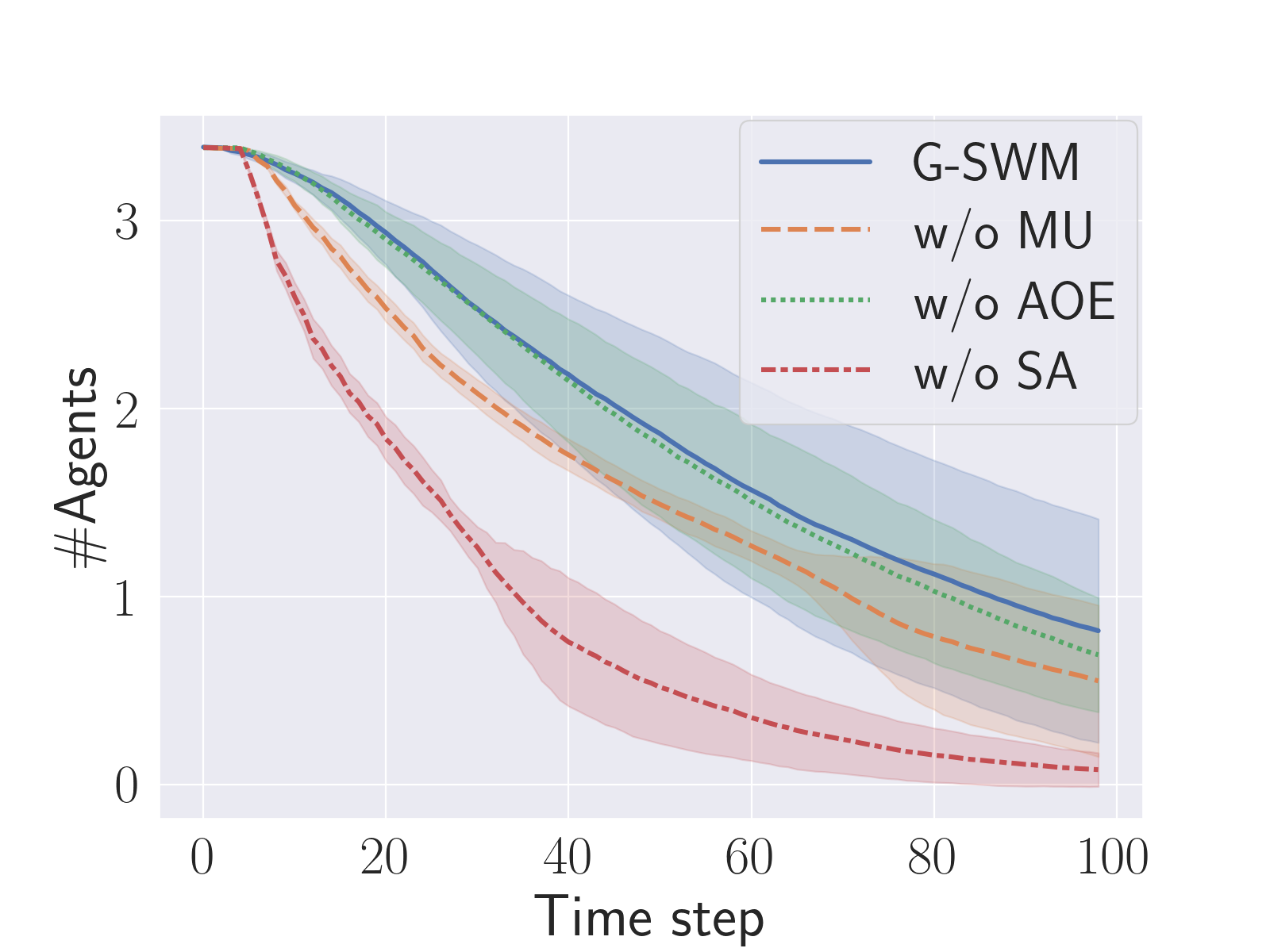}
    %   \caption{Hi}
    \end{subfigure}%
    \hspace{8mm}
    \begin{subfigure}[b]{0.35\linewidth}
      \centering
      \includegraphics[width=\linewidth]{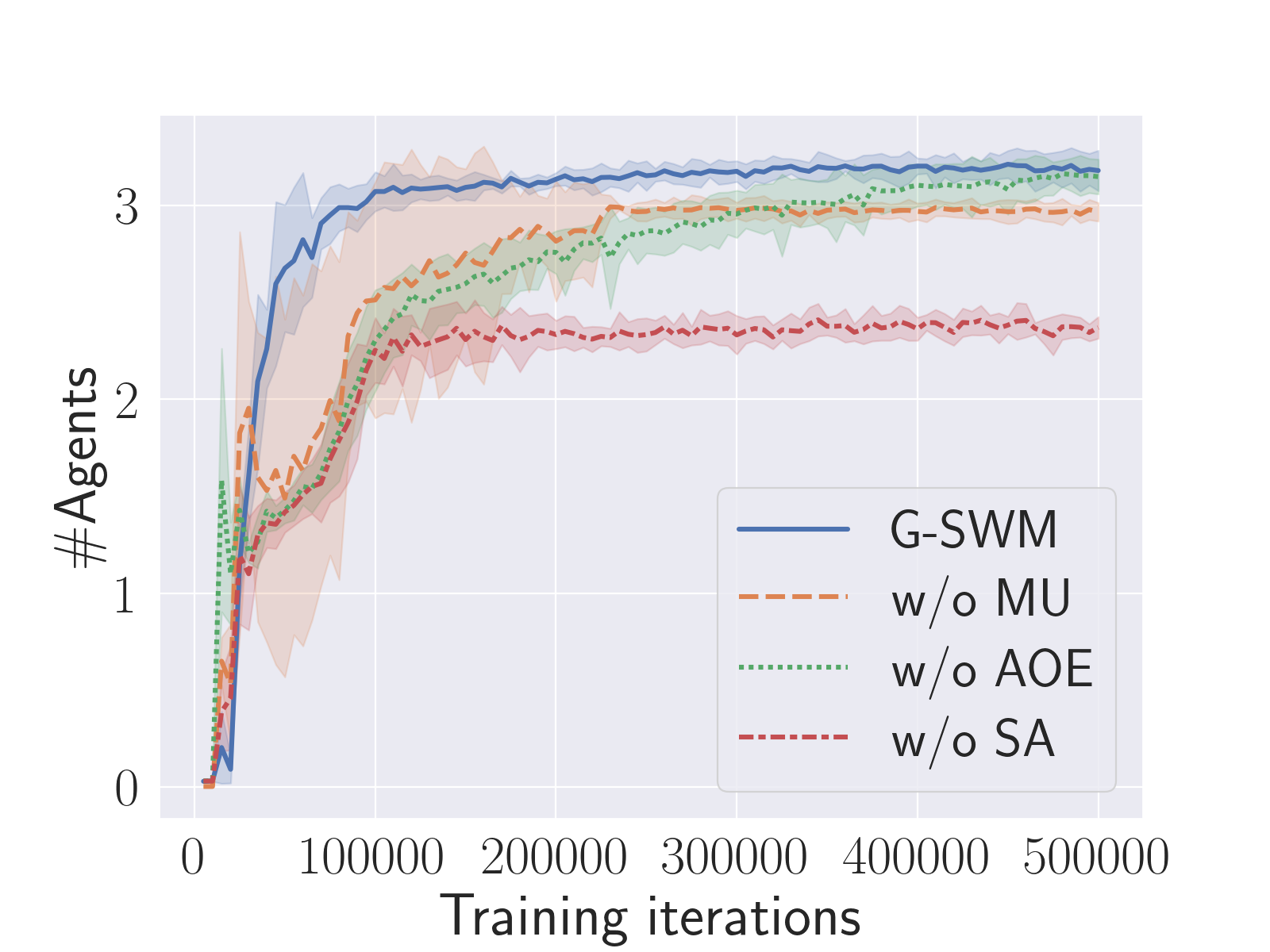}
    %   \caption{Hi}
    \end{subfigure}
    \vspace{-3mm}
    \caption{Quantitative results of the ablation study. The left plot shows the number of agents that stay within the maze corridors over 100 timesteps. The right plot shows the convergence curve of this metric during training, averaged over the first 10 generation steps. Each experiment is run with 5 random seeds. 
    %\red{Make all text in the figures much larger.} Done
    }
    \label{fig:maze_num}
\end{figure*}

As an ablation study, we introduce several variants of \ours by respectively removing the multimodal uncertainty (MU) modeling (i.e., removing $\bz^{\state}$), the attention on environment (AOE) mechanism, and the modeling of situation awareness (SA). We count the number of agents that correctly stay within the corridors of the maze over time as an indicator of generation quality. Figure~\ref{fig:maze_num} shows the results over 100 timesteps and the convergence curve of this metric during training. Without $\bz^{\state}$ or situation awareness, the model cannot correctly predict the agent behaviour in the maze and thus clearly underperforming. Without AOE but with the attention-less encoding of the full context, the model eventually reaches a similar performance as the complete model, but it takes much longer to converge as shown in Figure~\ref{fig:maze_num}. These results show that the proposed hierarchical modeling and the AOE mechanism can significantly improve generation performance when multimodal uncertainty and situation awareness are required.

% \begin{figure}[htbp]
%     \centering
%     \includegraphics[width=\linewidth]{figs/maze_num.png}
%     \caption{Number of agents that stay within the maze corridors over 100 timesteps. Each experiment is run with 5 seeds.}
%     \label{fig:maze_num}
% \end{figure}
% \begin{figure}[htbp]
%     \centering
%     \includegraphics[width=\linewidth]{figs/maze_convergence.png}
%     \caption{Number of agents that stay within the maze corridor during training. The numbers are averaged over the first 10 generated timesteps.}
%     \label{fig:maze_num_conv}
% \end{figure}

% However, we observe that this stochastic rollout is not stable in practice. In many cases, once the object deviates a little from the corridor due to noise in sampling, the trajectory will quickly become unpredictable. We believe this a fundamental limitation of modeling object dynamics with a simple RNN.

\subsection{3D Interactions}
\begin{figure*}[t!]
    \hspace{-8mm}
    \centering
    \includegraphics[width=0.95\linewidth]{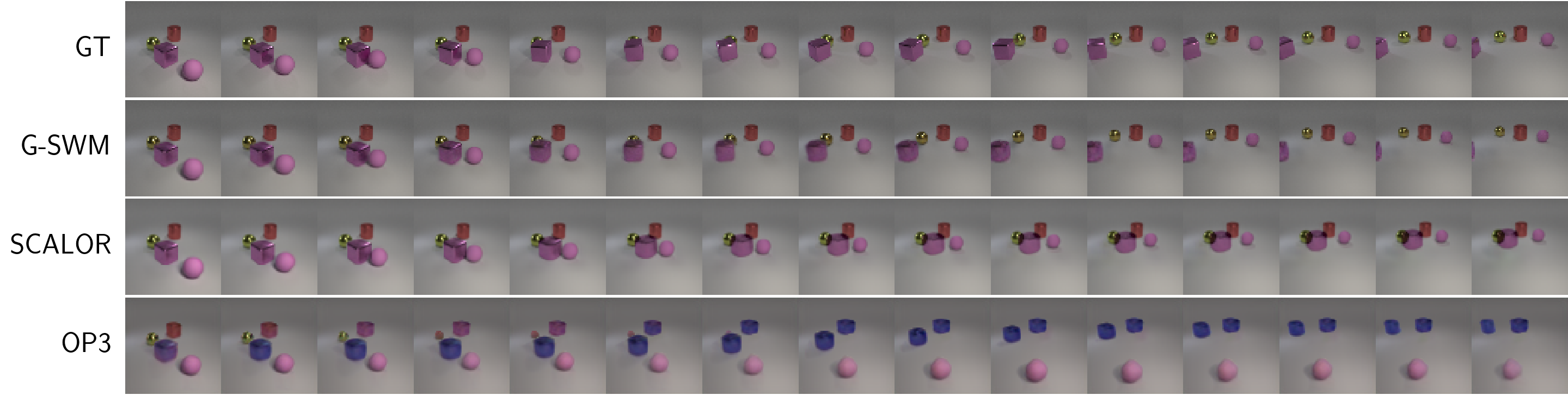}
    \caption{Generation results for the 3D realistic physics experiments. Given the first 10 frames of the episode, 90 frames are predicted (15 out of the first 30 generated frames are shown).}
    \label{fig:3d}
\end{figure*}

Though the above experiments clearly exhibit the different capabilities of our model, we also want to demonstrate how our model performs in a more realistic setting.
We use the CLEVR~\citep{clevr} dataset as a starting point to create a dynamic 3D environment. In this dataset, several 3D objects are placed on a surface and a ball moves towards the objects, knocking them around.
The shapes, positions, and colors of the objects are chosen randomly as well as the angle from which the ball enters the scene. For this dataset, we train on sequences of length 20. At test time, the model is given the first 10 ground truth frames, which are always before any interaction occurs. 

% We compare with OP3~\citep{op3}, as it provides interaction modeling while also being able to handle background.

% \begin{figure}[htbp]
%     \centering
%     \includegraphics[width=\linewidth]{figs/3d_compare.png}
%     \caption{Comparison of \ours, OP3, and SCALOR on the 3D dataset. 6 out of the first 12 generated frames are shown. \red{Show just 1 sample and move others to appendix. Move Fig 9 to Fig 8 and make the generation longer.}}
%     \label{fig:3d_compare}
% \end{figure}

Figure~\ref{fig:3d} shows the qualitative generation results on this dataset. More examples are given in Appendix. For comparison, we also test SCALOR and OP3 on this dataset.
% and the results are shown in Figure~\ref{fig:3d_compare}. 
Since OP3 assumes the Markov property of the generation process, it cannot even model the ball movement. Besides, with implicit object representation, OP3 produces very blurry images. SCALOR produces clean reconstructions of objects, but due to lack of interaction modeling, it cannot model collisions and different objects overlap when they collide.
%OP3 produces very blurry images and is very inaccurate in terms of prediction, while 
\ours can produce realistic generations even in this complex setting. Occlusions are correctly captured when one object is in front of another and the generated collisions are also realistic. In particular, the model predicts a collision event only when two objects are close in the 3D space (e.g., at the same depth), which further demonstrates \oursns's ability to handle interactions based on the depth information. Although the generated object appearances are similar to the ground truth, we do notice that the model fails to capture the purple cube in this example, instead of generating a cylinder. 
This may be because, in the ground truth, the cube spins when it is hit, which is difficult to model correctly. Nevertheless, our model is able to predict other changes to appearance such as the change in the objects size (in the 2D plane) as it moves further away from the camera.

%% file: tables/all_in_one.tex
% \begin{center}
\setlength{\tabcolsep}{4.5mm}{
\renewcommand{\arraystretch}{1.1}{
\begin{small}
\begin{tabular}{lccc}

\toprule
 & G-SWM & SCALOR & STOVE \\
\midrule
\textsc{Interaction} & $\mathbf{0.242 \pm 0.028}$ & $1.004 \pm 0.019$       & $0.411 \pm 0.007$       \\
\textsc{Occlusion}   & $\mathbf{0.072 \pm 0.015}$ & $0.272 \pm 0.019$       & $0.387 \pm 0.021$       \\
\textsc{2 Layer}     & $\mathbf{0.212 \pm 0.007}$ & $1.012 \pm 0.021$       & $-$                  \\
\textsc{2 Layer-D}   & $\mathbf{0.535 \pm 0.013}$ & $1.208 \pm 0.011$       & $-$                  \\

% \textsc{Interaction} & $\mathbf{0.2424 \pm 0.0276}$       & $1.0041 \pm 0.0194$       & $0.4108 \pm 0.0074$       \\
% \textsc{Occlusion}   & $\mathbf{0.0723 \pm 0.0146}$       & $0.2721 \pm 0.0186$       & $0.3865 \pm 0.0212$       \\
% \textsc{2 Layer} &  $\mathbf{0.2119 \pm 0.0068}$       & $1.0117 \pm 0.0208$       & $-$                  \\
% \textsc{2 Layer-D} & $\mathbf{0.5350 \pm 0.0128}$       & $1.2084 \pm 0.0108$       & $-$                  \\
\bottomrule
\end{tabular}
\end{small}
}
}
% \end{center}

%% file: 05_conclusion.tex
\section{Conclusion}

In this paper, we proposed a new object-centric temporal generative model for world modeling. In experiments, we performed an extensive investigation on the generation performance of previously proposed models and ours, which has been missing in the literature. Through this comparison, we demonstrated that \ours implementing all important abilities jointly in a single model achieves superior or comparable performance in all tasks. We also demonstrated that \ours successively achieves two new important abilities, multimodal uncertainty and situated behavior. A future direction is to resolve some weakness of the spatial-attention-based object models and to support the temporal discovery model. This would help apply this model to improve model-based reinforcement learning and planning.

%% file: supplementary.tex
\twocolumn[
\icmltitle{Supplementary Material for \\``Improving Generative Imagination in Object-Centric World Models''}
]
\appendix

\section{Model Details}
In this section, we will give a detailed description of each stage, especially those not described in detail in the main text.

For each timestep $t$, we will describe the generation of $\bz^\ctx_{t}, \tilbz_{t},\tilbo_{t}, \barbz_{t},  \barbo_{t}, \bo_{t}, \bx_{t}$ (in that order), given the full history $\bz^\ctx_{<t}$, $\dotbz_{<t}$, and $\bo_{<t}$. Generation consists of the following stages:
\begin{enumerate}
    \item \textbf{Context}. Given context history $\bz^\ctx_{<t}$, we generate the new context $\bz^\ctx_{t}$. 
    \item \textbf{Propagation}. We compute $\{\tilbz^{k}_{t}\}_{k=1}^{K}$, and then update the object attributes $\{\bo_{t-1}^{k}\}_{k=1}^K$  to $\{\tilbo^{k}_{t}\}_{k=1}^K$. 
   \item \textbf{Discovery}. A grid of $H\times W$ new object latents $\{\barbz^{ij}_{t}, (i, j)\in\{(1, 1), \ldots, (H, W)\}\}$ will be sampled from some predefined prior, and then for each $(i, j) \in \{(1, 1),\ldots, (H, W)\}$, $\barbo^{ij}_{t}$ will be obtained by passing each $\barbz^{ij}_{t}$ through some deterministic function. As mentioned in the main text, discovery will only be used during inference but not generation. Here, the discovery priors are only used to regularize inference. 
 
   \item \textbf{Rendering}. Given the set of propagated objects $\tilbo_{t}$ and discovered objects $\barbo_{t}$, we will select a maximum number of $K$ objects $\{\bo^k\}_{k=1}^K$ with the highest presence value. These objects will also be propagated to the next timestep. We then render the frame $\bx_{t}$ using the selected objects $\{\bo^k\}_{k=1}^K$, which generates the foreground image $\bmu^{\fg}_{t}$ and mask $\bal_t$, and the context latent  $\bz^\ctx_{t}$, which generates the background image $\bmu^{\bgr}_{t}$.
\end{enumerate}
% \icmltitle{Learning and Simulation in Generative Structured World Models}
% % \icmltitle{Simulating the World via Generative Structured World Models}
Below we describe the implementation details of each stage. 
\subsection{Context}

\textbf{Generation}. The prior $p_\theta(\bz^{\ctx}_t|\bz^{\ctx}_{<t})$ is implemented as follows:
\begin{align}
    \bh^\ctx_{t} &= \RNN^{\ctx}_{\prior}(\bz^\ctx_{t-1}, \bh^\ctx_{t-1})\\{}
    [\bmu^\ctx_{t}, \bsig^\ctx_{t}] &= \MLP^{\ctx}_{\prior}(\bh^\ctx_{t})\\
    \bz^\ctx_{t} &\sim \mathcal N(\bmu^\ctx_{t}, \bsig^\ctx_{t}).
\end{align}
\textbf{Inference}. The posterior $q_\phi(\bz^{\ctx}_t|\bx_t, \bz^{\ctx}_{<t})$ is implemented as follows:
\begin{align}
    \hatbh^\ctx_{t} &= \RNN^{\ctx}_{\post}(\bz^\ctx_{t-1}, \hatbh^\ctx_{t-1})\\{}
    \bee^{\ctx}_{\enc, t} &= \Conv^{\ctx}_{\enc}(\bx_t)\\{}
    [\bmu^\ctx_{t}, \bsig^\ctx_{t}] &= \MLP^{\ctx}_{\post}([\hatbh^\ctx_{t}, \bee^{\ctx}_{\enc, t}])\\
    \bz^\ctx_{t} &\sim \mathcal N(\bmu^\ctx_{t}, \bsig^\ctx_{t}).
\end{align}
\subsection{Propagation}
\textbf{Generation}. The overall procedure is described in the main text, so we only describe some network implementation details. 

The self-interaction encoding $\bee^{k,k}_t$, pairwise-interaction encoding $\bee^{k,j}_t$, and the interaction weights $w^{k,j}_t$ are computed as follows:
\eq{
    \bee^{k,k}_t &= \MLP^{\self}_\prior(\bu^k_t)\\
    \bee^{k,j}_t &= \MLP^{\rel}_\prior(\bu^k_t, \bu^j_t)\\
    w^{k,j}_t &= \MLP^{\weight}_\prior(\bu^k_t, \bu^j_t).
}
Given the hidden state $\bh^k_t$ of the OS-RNN, the state latent $\tilbz^{\state,k}_t$ is computed as follows:
\eq{
    [{\tilde \bmu}^{\state,k}_t, {\tilde \bsig}^{\state,k}_t] &= \MLP^{\state}_{\prior}(\bh_t^k)\ \\{}
    \tilbz^{\state,k}_t&\sim \mathcal N({\tilde \bmu}^{\state,k}_t, {\tilde \bsig}^{\state,k}_t),
}
and given the state latent $\bz^{\state,k}_t$, the attribute latents $\bz^{\att,k}_t = [\bz^{\pres, k}_t, \bz^{\depth, k}_t, \bz^{\where, k}_t, \bz^{\what, k}_t]$ are computed as follows:
\eq{
    [\tilde \brho^{\pres, k}, {\tilde \bmu}^{\depth,k}_t, {\tilde \bsig}^{\depth,k}_t,& {\tilde \bmu}^{\where,k}_t, {\tilde \bsig}^{\where,k}_t\nonumber \\{} 
    {\tilde \bmu}^{\what,k}_t, {\tilde \bsig}^{\what,k}_t] &= \MLP^{\att}_{\prior}(\tilbz^{\state, k}_t)\ \\{}
    \tilbz^{\pres,k}_t&\sim \text{Bernoulli}(\tilde \brho^{\pres, k})\\
    \tilbz^{\depth,k}_t&\sim \mathcal N({\tilde \bmu}^{\depth,k}_t, {\tilde \bsig}^{\depth,k}_t)\\
    \tilbz^{\where,k}_t&\sim \mathcal N({\tilde \bmu}^{\where,k}_t, {\tilde \bsig}^{\where,k}_t)\\
    \tilbz^{\what,k}_t&\sim \mathcal N({\tilde \bmu}^{\what,k}_t, {\tilde \bsig}^{\what,k}_t).
}

\textbf{Inference}. We only need to describe the implementation of $q_\phi(\tilbz^{\state, k}_t|\bx_t, \bz^{\ctx}_{<t}, \dotbz_{<t})$. First, a posterior OS-RNN will be used to update the posterior object state:
\eq{
  \hatbh^{k}_t = \text{RNN}^\os_{\post}([\bo^k_{t-1},\bz^{\state,k}_t, \bee^{\ctx, k}_{t-1}, \bee^{\rel, k}_{t-1}], \hatbh^{k}_{t-1})\ . 
 }
Here, $\bee^{\ctx, k}_{t-1}$ is computed using exactly the same process and network as generation, and $\bee^{\rel, k}_{t-1}$ is computed using a similar process during generation but with a separate set of posterior networks $\MLP^{\self}_{\post}$, $\MLP^{\rel}_{\post}$, and $\MLP^{\weight}_{\post}$.

Then, a proposal region of the image $\bx_t$ centered at the previous object location $\bo^{xy, k}_{t-1}$ is extracted and encoded. The size $\bs^{\prop}_t$ (2-dimensional for $(h, w)$) of this proposal area is computed from $\hatbh^{k}_t$:
\begin{equation}
    \bs^{\prop}_t = \bo^{hw,k}_{t-1} + s^{\min} + (s^{\max} - s^{\min})\cdot \sigma(\MLP^{\prop}(\hatbh^{k}_t))
\end{equation}
Where $s^{\min}$ and $s^{\max}$ are hyperparameters that control the minimum and maximum proposal update size. After that, the proposal is extracted and encoded:
\begin{align}
    \bg^{\prop, k}_{t} &= \STransform(\bx_{t}, \bo^{xy, k}_{t-1}, \bs^{\prop}_t)\\
    \bee^{\prop, k}_{t} &= \Conv^{\prop}(\bg^{\prop, k}_{t}).
\end{align}
Then $\hatbh^{k}_t$ and $\bee^{\prop, k}_t$ will be used to infer $\tilbz^{\state, k}_t$:
\begin{align}
    [{\tilde \bmu}^{\state,k}_t, {\tilde \bsig}^{\state,k}_t] &= \MLP^{\state}_{\post}([\hatbh_t^k, \bee^{\prop, k}_t])\ ,\\{}
    \tilbz^{\state,k}_t&\sim \mathcal N({\tilde \bmu}^{\state,k}_t, {\tilde \bsig}^{\state,k}_t).
\end{align}
\textbf{Attribute updates}. For this part we describe the details of object attribute update function $f^{\pres}$, $f^{\depth}$, $f^{\where}$, and $f^{\what}$. These functions are implemented as follows:
\begin{align}
[\bg^{\depth, k}_t, &\bg^{\where, k}_t, \bg^{\what, k}_t] = \sigma(\MLP^{\gate}(\tilbz^{\state, k}_t))\\
    \tilbo^{\pres, k} &= \bo^{\pres, k}_{t-1} \cdot \tilbz^{\pres}_t\\
    \tilbo^{\depth, k}_t &= \bo^{\depth, k}_t + c^{\depth}\cdot \bg^{\depth, k}_t \cdot \tilbz^{\depth, k}_t\\
    \tilbo^{xy, k}_t &= \bo^{xy, k}_t + c^{xy}\cdot \bg^{xy, k}_t \cdot \tanh(\tilbz^{xy, k}_t)\\
    \tilbo^{hw, k}_t &= \bo^{hw, k}_t + c^{hw}\cdot \bg^{hw, k}_t \cdot \tanh(\tilbz^{hw, k}_t)\\
    \tilbo^{\what, k}_t &= \bo^{\what, k}_t + c^{\what}\cdot \bg^{\what, k}_t \cdot \tanh(\tilbz^{\what, k}_t).
\end{align}
Note we split $\bo^{\where}$ into $\bo^{hw}$ and $\bo^{xy}$. Here, $c^{\depth}, c^{xy}, c^{hw}, c^{\what}$ are real-valued hyperparameters between $0$ and $1$ that control the degree of update we want.  Note that for $f^{\depth}$, $f^{\where}$, and $f^{\what}$, the corresponding update gates $\bg^{\depth, k}_t$, $\bg^{\where, k}_t$, and $\bg^{\what, k}_t$ will first be computed from $\tilbz^{\state, k}_t$ and used to mask the update values.

\subsection{Discovery}
\textbf{Generation}. We assume an independent prior for each object:
\eq{
    p(\barbz^{ij}_{t}) = p(\barbz^{\state, ij}_{t})p(\barbz^{\pres, ij}_{t}) 
    \Big\{p(\barbz^{\depth, ij}_{t})\nonumber\\
    p(\barbz^{\where, ij}_{t})p(\barbz^{\what, ij}_{t})\Big\}^{\barbz^{\pres, ij}}.
}
All of these priors are fixed Gaussian distributions with chosen mean and variance except for $p(\barbz^{\pres})$, which is a Bernoulli distribution.

\textbf{Inference}. We feed in the image $\bx_t$ along with the difference between the $\bx_t$ and the reconstructed background into an encoder to get an encoding of the current image $\bee^{\img}_t$ of shape $(H, W, C)$:
\eq{
    \bee^{\img}_t = \Conv^{\disc}([\bx_t, \bx_t - \bmu^{\bgr}_t])
}
To infer $\barbz_t$, besides the current image $\bx_t$, we also consider the propagated objects $\{\tilbo^k_t\}_{k=1}^K$ to prevent rediscovering already propagated objects. We adopt the same mechanism in SILOT to condition discovery on propagation. Specifically, for each discovery cell $(i, j) \in \{(1,1), \ldots, (H, W)\}$, a vector $\bee^{\cond, ij}_t$ will be computed as a weighted sum of all propagated objects $\{\tilbz^{k}_t\}_{k=1}^K$, with the weights computed by passing the relative distance between the propagated object $\tilbo^{xy, k}_t$ and the cell center $\bc^{ij}$ into a Gaussian kernel:
\begin{equation}
    \bee^{\cond, ij}_t = \sum_{k=1}^K G(\tilbo^{xy, k} - \bc^{ij}, \sigma^{\cond})\cdot \MLP^{\cond}(\tilbo^{k}_t)
\end{equation}
where $G$ is a 2-D Gaussian kernel, and $\sigma^{\cond}$ is a hyperparameter.

The discovered latents will then be computed conditioned on the image features and the encoding of propagated objects:
\begin{align}
    [\bar\bmu^{\state, ij}_t, \bar\bsig^{\state, ij}_t, &\bar \brho^{\pres, ij},{\bar \bmu}^{\depth,ij}_t, {\bar \bsig}^{\depth,ij}_t, {\bar \bmu}^{\where,ij}_t, \nonumber \\{}{\bar \bsig}^{\where,ij}_t, 
    {\bar \bmu}^{\what,ij}_t, &{\bar \bsig}^{\what,ij}_t] = \MLP^{\disc}([\bee^{\img, ij}_t, \bee^{\cond, ij}_t]\\
    \barbz^{\state, ij}_t &\sim \mathcal N(\bar\bmu^{\state, ij}_t, \bar\bsig^{\state, ij}_t)\\
        \barbz^{\pres,ij}_t&\sim \text{Bernoulli}(\bar \brho^{\pres, ij})\\
    \barbz^{\depth,ij}_t&\sim \mathcal N({\bar \bmu}^{\depth,ij}_t, {\bar \bsig}^{\depth,ij}_t)\\
    \barbz^{\where,ij}_t&\sim \mathcal N({\bar \bmu}^{\where,ij}_t, {\bar \bsig}^{\where,ij}_t)\\
    \barbz^{\what,ij}_t&\sim \mathcal N({\bar \bmu}^{\what,ij}_t, {\bar \bsig}^{\what,ij}_t)
\end{align}
Finally, we compute the object representation $\barbo^{ij}_t$ using these latents. For $\barbo^{\pres, ij}_{t}, \barbo^{\depth, ij}_{t}, \barbo^{\what, ij}_{t}$, they will just be equal to $\barbz^{\pres, ij}_{t}, \barbz^{\depth, ij}_{t}, \barbz^{\what, ij}_{t}$. For $\barbo^{\where, ij}_t = [\barbo^{hw, ij}_t, \barbo^{xy, ij}_t]$, we want $\barbo^{hw, ij}_t$ to be in range $(0, 1)$ and $\barbo^{xy, ij}_t$ in range $(-1, 1)$. Besides, as in SILOT, $\barbz^{xy, ij}_t$ is relative to the cell center $\bc^{ij}$, so we need to transform relative locations to global locations using
\begin{align}
    \barbo^{hw, ij}_t &= \sigma(\barbz^{hw, ij}_t)\\
    \barbo^{xy, ij}_t &= \bc^{ij} + 2\cdot \tanh(\barbz^{xy, ij}_t) / [W, H]
\end{align}
where $\bc^{ij} = 2\cdot ( [i, j] + 0.5) / [W, H] - 1$
\subsection{Rendering}
The background image $\bmu^{\bgr}_t$ will be decoded from the context latent $\bz^{\ctx}_t$:
\eq{
    \bmu^{\bgr}_t = \Deconv^{\ctx}_{\dec}(\bz^{\ctx}_t)
}
For foreground, we will first select a set of $K$ objects $\{\bo^k_t\}_{k=1}^K$ from the set of discovered and propagated objects $\barbo_t\cup \tilbo_t$. To render the set of selected objects $\{\bo^{k}_t\}_{k=1}^K$ into the foreground image $\bmu^{\fg}_t$ and foreground mask $\bal_t$, a similar procedure in SILOT is used. First, individual object appearance $\hat \by^{\att, k}_t$ and mask $\hat \bal^{\att, k}_t$ are computed from $\bo^{\what, k}_t$ and $\bo^{\pres, k}_t$:
\begin{align}
    [\by^{\att, k}_t, \bal^{\att, k}_t] &= \sigma(\Deconv^{\what}(\bo^{\what, k}_t))\\
    \hat \bal^{\att, k}_t &= \bal^{\att, k}_t \cdot \bo^{\pres, k}_t\\
    \hat \by^{\att, k}_t &= \hat \bal^{\att, k}_t\cdot \by^{\att, k}. 
\end{align}
Here, $\hat \by^{\att, k}_t$ and $\hat \bal^{\att, k}_t$ will be of a small glimpse size $(H_{g}, W_{g})$. We will then transform them into full image size $(H_{\img}, W_{\img})$ by putting them in an empty canvas using a (inverse) Spatial Transformer:
\begin{align}
    \by^{k}_t = \STransform^{-1}(\hat \by^{\att, k}_t, \bo^{\where, k}_t)\\
    \bal^{k}_t = \STransform^{-1}(\hat \bal^{\att, k}_t, \bo^{\where, k}_t)
\end{align}
Then $\bmu^{\fg}_t$ and $\bal_t$ will be computed as pixel-wise weighted sums of these image-sized maps:
\begin{align}
    \bw^k_t &= \frac{\bal^{k}_t \cdot \sigma(\bo^{\depth, k}_t)}{\sum_{j=1}^K \bal^{j}_t \cdot \sigma(\bo^{\depth, j}_t)}\\
    \bmu^{\fg}_t &= \sum_{k=1}^K \bw^k_t \cdot \by^k_t\\
    \bal_t &= \sum_{k=1}^K \bw^k_t \cdot \bal^k_t
\end{align}
The final rendered image will be $\bmu_t = \bmu^{\fg}_t + (1 - \bal_t) \bmu^{\bgr}_t$. The likelihood $p_\theta(\bx_t | \bz^{\ctx}_{\le t}, \dotbz_{\le t})$ is then
\begin{equation}
    p_\theta(\bx_t | \bz^{\ctx}_{\le t}, \dotbz_{\le t}) = \mathcal N(\bx_t| \bmu_t, \sigma^2\bI)
\end{equation}
where $\sigma$ is a hyperparameter.

\section{Architectures, Hyperparameters, and Training}

\subsection{Training}

For all experiments, we use the Adam~\citep{adamoptim} optimizer with a learning rate of $1\times 10^{-4}$ except for the maze dataset. We use a batch size of $16$ for all experiments. Gradient clipping~\citep{gradclip} with a maximum norm of $1.0$ is applied. For both $\barbz^{\pres, ij}_k$ and $\tilbz^{\pres, k}_t$, we use a Gumbel-Softmax relaxation~\citep{gumbel} with temperature $\tau$ to make sampling differentiable.

For experiments on datasets without background, we manually set $\bmu^{\bgr}_t$ to empty images. For the maze dataset,  we turn off the gradient of the foreground module and only learn to reconstruct background for the first $500$ steps. Also, we use a learning rate of $5\times 10^{-5}$ instead of $1\times 10^{-4}$.

\subsection{Architectures}

All RNNs are implemented as LSTMs~\citep{lstm}. For all equations that describe RNN recurrence, the notation $\bh$ includes both the hidden state and cell state used in common LSTMs. However, when $\bh$ is used as an input to another network, we use only the hidden state. For all initial states ($\bh_0$), we treat them as learnable parameters with unit Gaussian random initialization. For both the prior and posterior object-state RNN, inputs are first embedded with a single fully connected layer denoted by $\MLP^{\os}_{\prior}$ and $\MLP^{\os}_{\post}$.

For all networks that output variances of Gaussian distributions, we apply a softplus function to ensure that the variances are positive. For all networks that output the parameters of Bernoulli distributions (for $\bz^{\pres}$), we apply a sigmoid function.

Table~\ref{tab:arch} lists all networks. Here, $\LSTM(a, b)$ denotes an LSTM with input size $a$ and hidden size $b$. For MLPs, the Architecture column lists the hidden layer sizes, not including input and output layer.  The identity of input and output variables can be found in equations where each network appears, and the dimensions of these variables will be given in Section~\ref{sec:hyper}. 

For all network layers except for output layers, we use the CELU \citep{celu} activation function. For all convolution layers except for output layers, we use group normalization \citep{groupnorm} with 16 channels per group. Note that $\MLP^{\state}_{\prior}, \MLP^{\state}_{\post}, \MLP^{\att}_{\prior}, \MLP^{\gate}$ are implement as stride-1 convolutions to facilitate parallel computation. 

In Table~\ref{tab:arch}, $\Conv^{\disc}$ is implemented with ResNet18 \citep{resnet} by taking the feature volume from the third block ($1/8$ of the image size) and  applying a stride-1 or -2, $3\times3$ convolution layer depending on the grid size $(H, W)$ (in this work $H=W$ and is either $8$ or $4$) to obtain $\bee^{\img}_t$. Table~\ref{tab:env_encoder}, Table~\ref{tab:env_decoder}, Table~\ref{tab:env_att_encoder}, Table~\ref{tab:prop_encoder}, and Table~\ref{tab:what_decoder} list other convolutional encoders and decoders that are referred to in Table~\ref{tab:arch}. In these tables, $\text{Subconv}$ denotes a sub-pixel convolution \citep{subpixel} implemented by a normal convolution layer plus a PyTorch $\text{PixelShuffle}$ operation. The stride of $\text{Subconv}$ will be used as a parameter for $\text{PixelShuffle}$. GN($n$) denotes group normalization with $n$ groups.

\begin{table*}[htbp]
    \caption{Network details}
    \centering
    \input{tables/arch}
    \label{tab:arch}
\end{table*}

\begin{table}[htbp]
    \caption{$\Conv^{\ctx}$}
    \centering
    \input{tables/networks/env_encoder}
    \label{tab:env_encoder}
\end{table}
\begin{table}[htbp]
    \caption{$\Deconv^{\ctx}$}
    \centering
    \input{tables/networks/env_decoder}
    \label{tab:env_decoder}
\end{table}
\begin{table}[htbp]
    \centering
    \caption{$\Conv^{\ctx}_{\att}$}
    \input{tables/networks/env_att_encoder}
    \label{tab:env_att_encoder}
\end{table}
\begin{table}[htbp]
    \centering
    \caption{$\Conv^{\prop}$}
    \input{tables/networks/prop_encoder}
    \label{tab:prop_encoder}
\end{table}
\begin{table}[htbp]
    \centering
    \caption{$\Deconv^{\what}$}
    \input{tables/networks/what_decoder}
    \label{tab:what_decoder}
\end{table}

\subsection{Hyperparameters}
\label{sec:hyper}
Table~\ref{tab:hyper} lists the hyperparameters for the \textsc{2 Layer} dataset. Hyperparameters for other experiments are similar.

\begin{table}[htbp]
\caption{Hyperparameters}
    \centering
    \input{tables/hyper}
    \label{tab:hyper}
\end{table}

\section{Dataset Details}
\subsection{Bouncing Balls}

In all settings, the balls bounce off the walls of the frame, and no new balls are introduced in the middle of an episode. Each episode has a length of 100. We split our data into 10,000 episodes for the training set, and 200 episodes each for the validation set and test set.

In both the \textsc{Occlusion} and \textsc{Interaction} settings, there are 3 balls each with a color drawn from a set of 5 colors (blue, red, yellow, fuchsia, aqua), but for the \textsc{Occlusion} case we do not allow duplicate colors.

\subsection{Random Single Ball}

In this dataset, a single ball moves down the center of the frame for 9 timesteps. After 5 timesteps, the ball randomly changes direction and moves towards either the bottom left corner or the bottom right corner for the remaining 4 timesteps. We split our data into 10,000 episodes for the training set, and 100 episodes each for the validation set and test set.

\subsection{Maze}

The mazes are created using the \texttt{mazelib} library\footnote{\url{https://github.com/theJollySin/mazelib}} and then removing dead ends manually. For the first frame, 3 or 4 agents of a random color drawn from 6 colors (red, lime, blue, yellow, cyan, magenta) are randomly placed in the corridors. The agents only move within the corridors and continue in a straight path until it reaches an intersection. It then randomly chooses a path, each with equal probability. Each episode has a sequence length of 99. We split our data into 10,000 episodes for the training set, and 100 episodes each for the validation set and test set.

\subsection{3D Interactions}

We generate the 3D Interactions dataset using Blender \citep{blender}, with the same base scene and object properties as the CLEVR dataset \citep{clevr}. In this dataset, we split our dataset into 2920 episodes for training, and 200 episodes for validation and test. Each episode has a length of 100.

We use three different objects (sphere, cylinder, cube), two different materials (rubber, metal), three different sizes, and five different colors (pink, red, blue, green, yellow) to generate the scenes. All objects move on a smooth surface without friction. 

To generate the dataset, we randomly put 3 to 5 objects in the camera scene, and launch a sphere into the scene colliding with other objects. The appearance and incident angle of this initial sphere are also randomly selected.

%To be specific, we limit the incident angle to the lower part of the scene.

\section{Experiment Details}

For all experiments that require generation, we set $\tilbz^{\pres, k}_t$ to $1$ for all timesteps at test time to ensure that objects do not disappear. Besides, we turn off discovery after the first timestep. For the bouncing ball experiments, during generation, we directly take the mean of each latent instead of sampling for all models since no stochasticity is involved. 

\subsection{Bouncing Balls}

We draw random sequences of length 20 for training. During testing, for each sequence of length 100, we condition on the first 10 frames and generate the following. We use 5 random seeds to run the experiments per model per dataset. All models were trained till full convergence and the results are computed using the model checkpoints that achieve the best performances on the validation set. For quantitative results, G-SWM is trained for 160000 steps for the \textsc{Interaction}, \textsc{Occlusion}, and \textsc{2 Layer} settings, and 120000 steps for the \textsc{2 Layer-D} settings.

\subsection{Random Single Ball}

We use full sequences of length 9 for training. At test time, each model is provided the first 5 timesteps of the ground truth, before the ball changes direction, and predicts the final 4 timesteps.

\subsection{Maze}

We use sequences of length 10 for training. During testing, we provide 5 ground truth timesteps as input. For quantitative results, G-SWM, including its variants, are trained for a maximum of 500000 steps.

\subsection{3D Interactions}

For this dataset, we use sequences of length 20 for training. However, since most interactions end after 30 steps, we draw training sequences only from the first 30 steps. During testing, for each test sequence of length 100, we provide the first 10 frames as input and generate the following frames. 

\section{Additional Results}

\begin{figure}[htbp]
    \centering
    \includegraphics[width=0.8\linewidth]{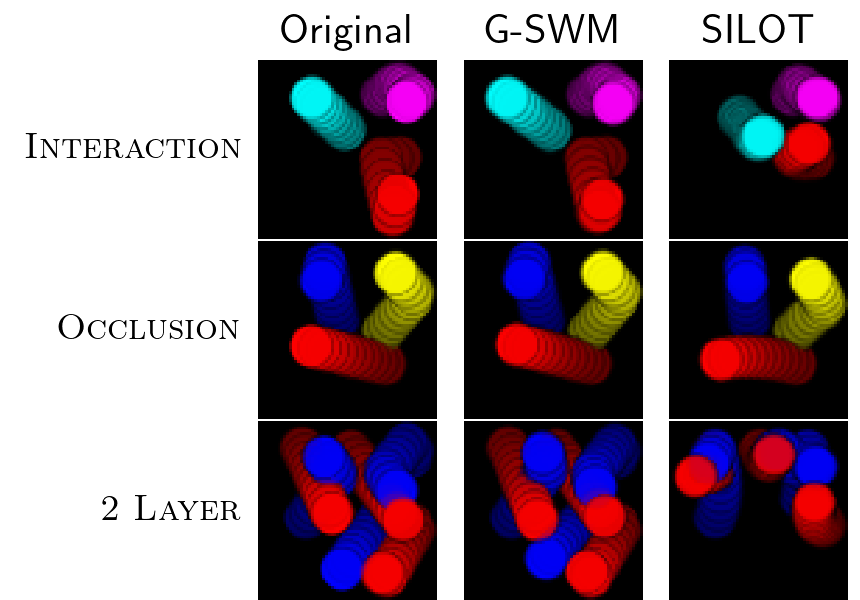}
    \caption{Generated frames of SILOT on the bouncing ball datasets.}
    \label{fig:silot}
\end{figure}

\textbf{SILOT.} We also test SILOT~\citep{silot} on the four bouncing ball datasets and the results are shown in Figure~\ref{fig:silot}. Being a very similar model to SCALOR, it can handle frequent occlusions and is scalable, but cannot handle the ball collisions well in the \textsc{Interaction}, \textsc{2 Layer}, and \textsc{2 Layer-D} settings, despite having a simple distance-based interaction module.

\textbf{Tracking Performance.} Table~\ref{tab:tracking} shows the tracking performance for the bouncing ball datasets. For tracking, we report the Multi-Object Tracking Accuracy (MOTA)~\citep{mot}, with an IoU threshold of 0.5.

\begin{table*}[t!bp]
\caption{Tracking performance on the bouncing ball datasets.}
\vspace{0.1in}
\label{tab:tracking}
\centering
\input{tables/tracking.tex}
\end{table*}

\begin{figure*}[htbp]
    \centering
    \includegraphics[width=\linewidth]{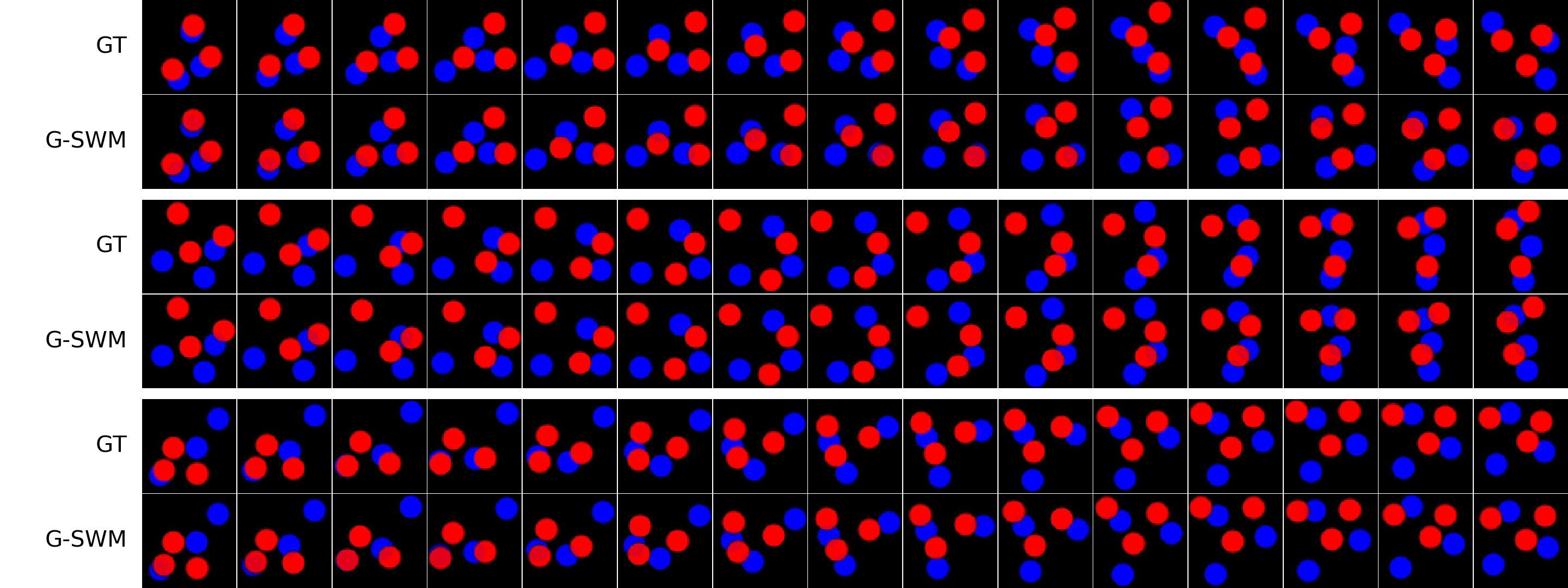}  
    \caption{\ours on the \textsc{2 Layer} dataset}
    \label{fig:two_layer_app}
\end{figure*}

\begin{figure*}[htbp]
    \centering
    \includegraphics[width=\linewidth]{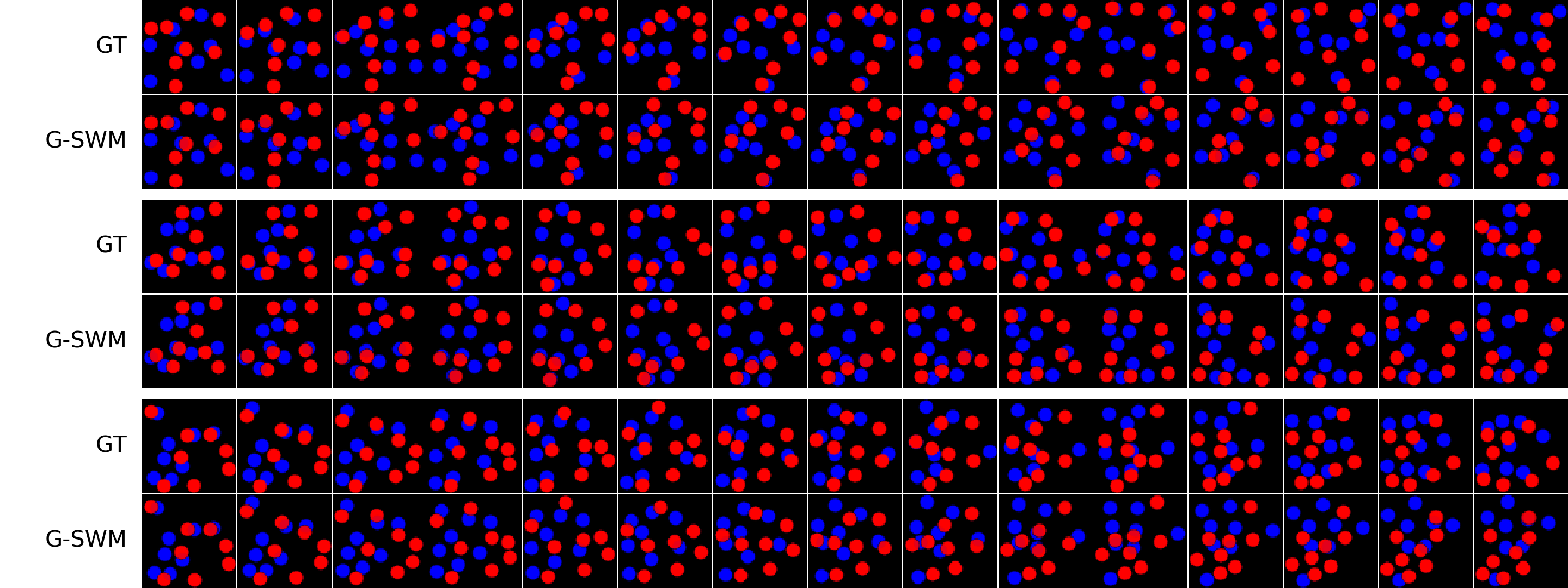}  
    \caption{\ours results on the \textsc{2 Layer-D dataset}}
    \label{fig:two_layer_dense_app}
\end{figure*}

\begin{figure*}[htbp]
    \centering
    \includegraphics[width=0.5\linewidth]{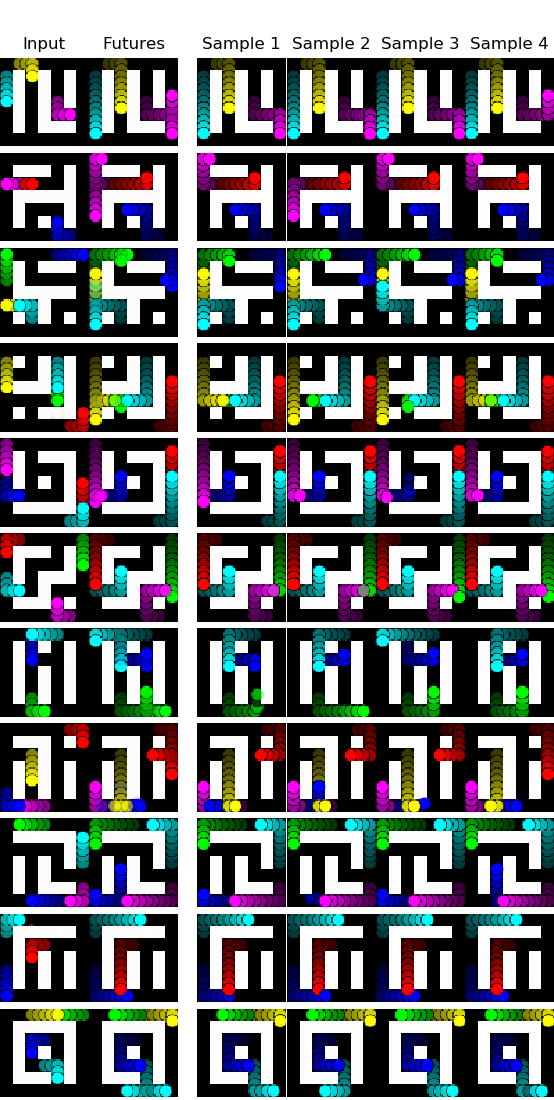}  
    \caption{Maze}
    \label{fig:maze_app}
\end{figure*}

\begin{figure*}[htbp]
    \centering
    \includegraphics[width=\linewidth]{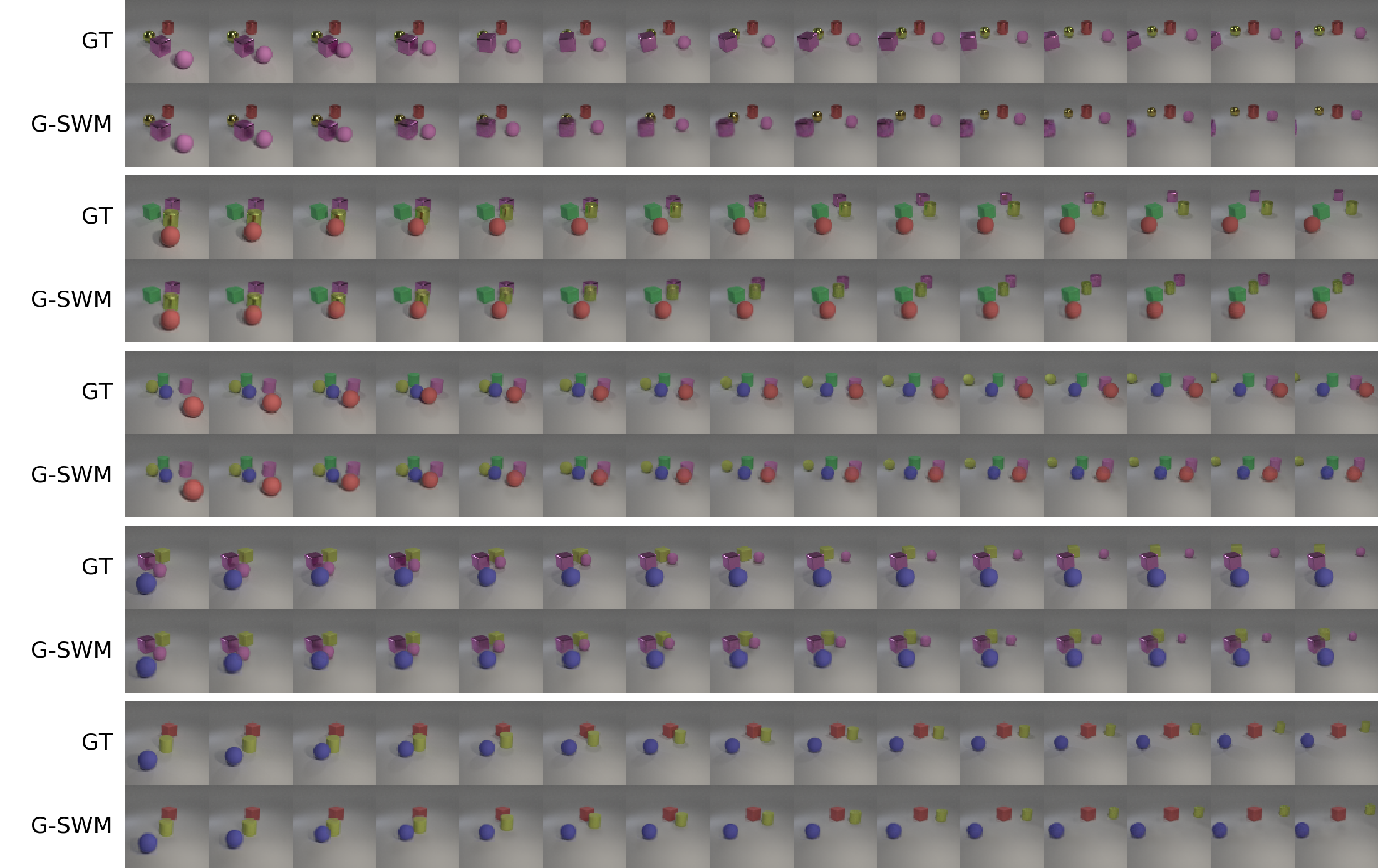}  
    \caption{\ours results on the 3D-Interactions dataset}
    \label{fig:3d_app}
\end{figure*}

\textbf{Additional Visualizations.} Figure~\ref{fig:two_layer_app} and Figure~\ref{fig:two_layer_dense_app} show visualizations of \ours on the two \textsc{2 Layer} datasets. Figure~\ref{fig:maze_app} and Figure~\ref{fig:3d_app} show additional results on the Maze and 3D datasets respectively. 

% For comparison we also trained SCALOR on the 3D dataset, and the generation is shown in Figure~\ref{fig:3d_scalor_app}. Note that SCALOR cannot simulate collision due to lack of a proper interaction module.

% \begin{figure*}[htbp]
%     \centering
%     \includegraphics[width=\linewidth]{figs/3d_scalor_appendix.png}  
%     \caption{SCALOR on 3D-Interactions}
%     \label{fig:3d_scalor_app}
% \end{figure*}

% \section{Baselines}

% \textbf{SILOT}.

% \textbf{SCALOR}.

% \textbf{STOVE}.

%% file: tables/arch.tex
\renewcommand{\arraystretch}{1.2}{
\begin{center}
% \begin{small}
\begin{tabular}{lll}
\toprule
 Description & Symbol & Architecture \\
\midrule
% Environment Generation
Context prior RNN & $\RNN^{\ctx}_{\prior}$ & LSTM(128, 128)\\
Generate $\bz^\ctx_t$ from $\bh^{\ctx}_t$ & $\MLP^{\ctx}_{\prior}$ &[128, 128]\\
Decode $\bz^\ctx_t$ into $\bmu^{\bgr}_t$ & $\Deconv^{\ctx}$ & See Table~\ref{tab:env_decoder}\\
% Context Inference
Context posterior RNN & $\RNN^{\ctx}_{\post}$ & LSTM(128, 128)\\
Infer $\bz^{\ctx}_t$  from $[\hatbh^{\ctx}, \bx_t]$& $\MLP^{\ctx}_{\post}$ & [128, 128]\\
Encode $\bx_t$ into $\bee^{\ctx}_{\enc}$ & $\Conv^{\ctx}$ & See Table~\ref{tab:env_encoder}\\
% Discovery
Encode $\bx_t$ into $\bee^{\img}_t$ & $\Conv^{\disc}$ & See the text\\
Encode $\tilbo^{k}_t$ during discovery& $\MLP^{\cond}$ & [128, 128]\\
Infer $\barbz^{ij}_k$ from $[\bee^{\img, ij}_t, \bee^{\cond, ij}_t]$ & $\MLP^{\disc}$ & [128, 128]\\
% Propagation Generation
Prior OS-RNN & $\RNN^{\os}_{\prior}$ & LSTM(128, 128)\\
OS-RNN input embedding & $\MLP^{\os}_{\prior}$ & []\\
% Interaction
Self-interaction encoding & $\MLP^{\self}_{\prior}$ & [128, 128]\\
Pairwise-interaction encoding & $\MLP^{\rel}_{\prior}$ & [128, 128]\\
Attention weights over object pairs & $\MLP^{\weight}_{\prior}$ & [128, 128]\\
% AOE
Attention on Environment encoder & $\Conv^{\ctx}_{\att}$ & See Table~\ref{tab:env_att_encoder}\\
% Predict attributes
Generate $\tilbz^{\state, k}_t$ from $\bh^k_{t}$ & $\MLP^{\state}_{\prior}$ & [128, 128]\\
Generate $\tilbz^{\att, k}_t$ from $\tilbz^{\state,k}_t$ & $\MLP^{\att}_{\prior}, \MLP^{\gate}$ & [128, 128]\\
Posterior OS-RNN & $\RNN^{\os}_{\post}$ & LSTM(128, 128)\\
OS-RNN input embedding & $\MLP^{\os}_{\post}$ & []\\
Predict proposal size $\bs^{\prop, k}_t$ & $\MLP^{\prop}$ & [128, 128]\\
Encode proposal into $\bee^{\prop, k}_t$ & $\Conv^{\prop}$ & See Table~\ref{tab:prop_encoder}\\
% Interaction
Self-interaction encoding & $\MLP^{\self}_{\post}$ & [128, 128]\\
Pairwise-interaction encoding & $\MLP^{\rel}_{\post}$ & [128, 128]\\
Attention weights over object pairs & $\MLP^{\weight}_{\post}$ & [128, 128]\\
Infer $\tilbz^{\state, k}_t$ from $[\hatbh^k_t, \bee^{\prop, k}_t]$ & $\MLP^{\state}_{\post}$ & [128, 128]\\
% Rendering
Decode $\bz^{\what, ij}_t$ into $\by^{\att, ij}_t, \bal^{\att, ij}_t$ & $\Deconv^{\what}$ & See Table~\ref{tab:what_decoder}\\

\bottomrule
\end{tabular}
% \end{small}
\end{center}
\vskip -0.1in
}

%% file: tables/networks/env_encoder.tex
\begin{tabular}{llll}
% \multicolumn{4}{l}{\Conv^{\env}}\\
\toprule
Layer           & Size/Ch. & Stride & Norm./Act.   \\ 
\midrule
Input           & 3       &        &              \\
Conv $7\times7$ & 64      & 2      & GN(4)/CELU \\
Conv $3\times3$ & 128      & 2      & GN(8)/CELU \\
Conv $3\times3$ & 256      & 2      & GN(16)/CELU \\
Conv $3\times3$ & 512      & 2      & GN(32)/CELU \\
Flatten & & \\
Linear & 128 &\\
\bottomrule
\end{tabular}

%% file: tables/networks/env_decoder.tex
\begin{tabular}{llll}
% \multicolumn{4}{l}{\Conv^{\env}}\\
\toprule
Layer           & Size/Ch. & Stride & Norm./Act.   \\ 
\midrule
Input           & 128 (1d)       &        &              \\
Reshape           & 128 (3d)       &        &              \\
Subconv $3\times3$ & 64      & 2      & GN(4)/CELU \\
Subconv $3\times3$ & 32      & 2      & GN(2)/CELU \\
Subconv $3\times3$ & 16      & 2      & GN(1)/CELU \\
Subconv $3\times3$ & 3      & 2      &  \\
Sigmoid &  &\\
\bottomrule
\end{tabular}

%% file: tables/networks/env_att_encoder.tex
\begin{tabular}{llll}
% \multicolumn{4}{l}{\Conv^{\env}}\\
\toprule
Layer           & Size/Ch. & Stride & Norm./Act.   \\ 
\midrule
Input           & 3       &        &              \\
Conv $3\times3$ & 16      & 2      & GN(1)/CELU \\
Conv $3\times3$ & 32     & 2      & GN(2)/CELU \\
Conv $3\times3$ & 64      & 2      & GN(4)/CELU \\
Conv $3\times3$ & 128      & 2      & GN(8)/CELU \\
Flatten & & \\
Linear & 128 &\\
\bottomrule
\end{tabular}

%% file: tables/networks/prop_encoder.tex
\begin{tabular}{llll}
% \multicolumn{4}{l}{\Conv^{\env}}\\
\toprule
Layer           & Size/Ch. & Stride & Norm./Act.   \\ 
\midrule
Input           & 3       &        &              \\
Conv $3\times3$ & 16      & 2      & GN(1)/CELU \\
Conv $3\times3$ & 32     & 2      & GN(2)/CELU \\
Conv $3\times3$ & 64      & 2      & GN(4)/CELU \\
Conv $3\times3$ & 128      & 2      & GN(8)/CELU \\
Flatten & & \\
Linear & 128 &\\
\bottomrule
\end{tabular}

%% file: tables/networks/what_decoder.tex
\begin{tabular}{llll}
% \multicolumn{4}{l}{\Conv^{\env}}\\
\toprule
Layer           & Size/Ch. & Stride & Norm./Act.   \\ 
\midrule
Input           & 128 (1d)       &        &              \\
Reshape           & 128 (3d)       &        &              \\
Subconv $3\times3$ & 64      & 2      & GN(4)/CELU \\
Subconv $3\times3$ & 32      & 2      & GN(2)/CELU \\
Subconv $3\times3$ & 16      & 2      & GN(1)/CELU \\
Subconv $3\times3$ & 3 + 1      & 2      & \\
Sigmoid &  &\\
\bottomrule
\end{tabular}

%% file: tables/hyper.tex
\begin{center}
\begin{small}
\begin{tabular}{lll}
\toprule
 Description & Symbol & Value \\
\midrule
% Dimensions
Image size & $(H_{\img}, W_{\img})$ & (64, 64)\\
Glimpse size & $(H_{g}, W_{g})$ & (16, 16)\\
Discovery grid size & $(H, W)$ & (4, 4)\\
Dimension of $\bz^{\pres, k}_t$ & & 1\\
Dimension of $\bz^{\depth, k}_t$ & & 1\\
Dimension of $\bz^{\where, k}_t$ & & 4\\
Dimension of $\bz^{\what, k}_t$ & & 64\\
Dimension of $\bz^{\state, k}_t$ & & 128\\
Dimension of $\bz^{\ctx}_t$ & & 128\\
Dimension of $\bee^{\img, ij}_t$ & & 128\\
Dimension of $\bee^{\cond, ij}_t$ & & 128\\
Dimension of $\bee^{\prop, k}_t$ & & 128\\
Dimension of $\bee^{k, k}_t$ & & 128\\
Dimension of $\bee^{k,j}_t$ & & 128\\
Dimension of $\bee^{\ctx}_{\enc, t}$ & & 128\\
Dimension of $\bee^{\ctx, k}_{t}$ & & 128\\

% General things
Training sequence length & $T$ & [2:20:2]\\
Curriculum milestones &  & [10k:90k:10k]\\
\#objects to select & $K$ & 10 \\

% Likelihood 
Likelihood variance & $\sigma$ & 0.2\\

% Background
AOE size & $s^{\ctx}$ & 0.25\\

% Tricks
Gaussian kernel sigma & $\sigma^{\cond}$& 0.1\\
Rejection IOU threshold & & 0.8 \\
Discovery dropout & & 0.5\\
Auxiliary KL parameter & $p$ & $1\times 10^{-10}$\\

% Priors
Gumbel-softmax temperature & $\tau$ & 1.0\\
$\barbz^{\pres, ij}_t$ prior & & $\text{Bern}(1\times 10^{-10})$\\
$\barbz^{\depth, ij}_t$ prior mean & & 0\\
$\barbz^{\depth, ij}_t$ prior stdev & & 1\\
$\barbz^{xy, ij}_t$ prior mean & & 0\\
$\barbz^{xy, ij}_t$ prior stdev & & 1\\
$\barbz^{hw, ij}_t$ prior mean & & -1.5\\
$\barbz^{hw, ij}_t$ prior stdev & & 0.3\\
$\barbz^{\what, ij}_t$ prior mean & & 0\\
$\barbz^{\what, ij}_t$ prior stdev & & 1\\

% Update scales
For updating $\tilbo^{\depth, k}_t$ & $c^{\depth}$& 1\\
For updating $\tilbo^{xy, k}_t$ & $c^{xy}$& 0.1\\
For updating $\tilbo^{hw, k}_t$ & $c^{hw}$& 0.3\\
For updating $\tilbo^{\what, k}_t$ & $c^{\what}$ & 0.2\\

% Proposal
Minimum proposal size & $s^{\min}$ & 0.0\\
Maximum proposal size & $s^{\max}$ & 0.2\\

\bottomrule
\end{tabular}
\end{small}
\end{center}
\vskip -0.1in

%% file: tables/tracking.tex
% \begin{center}
\setlength{\tabcolsep}{4.5mm}{
\renewcommand{\arraystretch}{1.1}{
\begin{tabular}{cccc}

\toprule
 & G-SWM & SCALOR & STOVE \\
\midrule
\textsc{Interaction}          & $0.9870 \pm 0.0032$     & $0.9688 \pm 0.0101$     & $\mathbf{0.9979 \pm 0.0005}$     \\
\textsc{Occlusion}            & $\mathbf{0.9919 \pm 0.0013}$     & $0.9447 \pm 0.0119$     & $0.9618 \pm 0.0023$     \\
\textsc{2 Layer}              & $\mathbf{0.9967 \pm 0.0041}$     & $0.9686 \pm 0.0102$     & $-$                  \\
\textsc{2 Layer-D}            & $\mathbf{0.9756 \pm 0.0066}$     & $0.9501 \pm 0.0087$     & $-$                  \\
\bottomrule
\end{tabular}
}
}
% \end{center}